\documentclass{article}

\usepackage[numbers, compress]{natbib}

\usepackage[preprint]{neurips_2019}

\usepackage[utf8]{inputenc} % allow utf-8 input
\usepackage[T1]{fontenc}    % use 8-bit T1 fonts
\usepackage{hyperref}       % hyperlinks
\usepackage{url}            % simple URL typesetting
\usepackage{booktabs}       % professional-quality tables
\usepackage{amsfonts}       % blackboard math symbols
\usepackage{nicefrac}       % compact symbols for 1/2, etc.
\usepackage{microtype}      % microtypography

\usepackage{amsmath}
\usepackage{graphicx}
\usepackage{xr}

\title{Unsupervised Emergence of Egocentric Spatial Structure from Sensorimotor Prediction}

\author{Alban Laflaqui\`ere \\
  AI Lab, SoftBank Robotics Europe \\
  Paris, France \\
  \texttt{alaflaquiere@softbankrobotics.com} \\
  \And
  Michael Garcia Ortiz \\
  AI Lab, SoftBank Robotics Europe \\
  Paris, France \\
  \texttt{mgarciaortiz@softbankrobotics.com} \\
}

\begin{document}

\maketitle

\begin{abstract}
Despite its omnipresence in robotics application, the nature of spatial knowledge and the mechanisms that underlie its emergence in autonomous agents are still poorly understood.
Recent theoretical works suggest that the Euclidean structure of space induces invariants in an agent's raw sensorimotor experience.
We hypothesize that capturing these invariants is beneficial for sensorimotor prediction and that, under certain exploratory conditions, a motor representation capturing the structure of the external space should emerge as a byproduct of learning to predict future sensory experiences.
We propose a simple sensorimotor predictive scheme, apply it to different agents and types of exploration, and evaluate the pertinence of these hypotheses.
We show that a naive agent can capture the topology and metric regularity of its sensor's position in an egocentric spatial frame without any a priori knowledge, nor extraneous supervision.
\end{abstract}

\section{Introduction}
\label{sec:Introduction}

Current model-free Reinforcement Learning (RL) approaches have proven to be very successful at solving difficult problems, but seem to lack the ability to extrapolate and transfer already acquired knowledge to new circumstances \cite{ cobbe2018quantifying, packer2018assessing}.
One way to overcome this limitation would be for learning agents to abstract from the data a model of the world that could support such extrapolation.
For agents acting in the world, such an acquired model should include a concept of \emph{space}, such that the spatial properties of the data they collect could be disentangled and extrapolated upon.

This problem naturally raises the question of the nature of space and how this abstract concept can be acquired.
This question has already been addressed philosophically by great minds of the past \cite{kant1781critique, Poincare1895, nicod1965geometrie}, among which the approach proposed by H.Poincar\'e is of particular interest, as it naturally lends itself to a mathematical formulation and concrete experimentation.
He was interested in understanding why we perceive ourselves as being immersed in a 3D and isotropic (Euclidean) space when our actual sensory experiences live in a multidimensional space of a different nature and structure (for instance, when the environment is projected on the flat heterogeneous surface of our retina).
He suggested that the concept of space emerges via the discovery of \emph{compensable sensory changes} that are generated by a change in the environment but can be canceled-out by a motor change.
This \emph{compensability} property applies specifically to \emph{displacements} of objects in the environment and of the sensor, but not to non-spatial changes (object changing color, agent changing its camera aperture...).
For instance, one can compensate the sensory change due to an object moving 1 meter away by moving 1 meter toward the object.
Moreover, this compensability property is invariant to the content of the environment, as the displacement of an object can be compensated by the same motor change regardless of the type of the object.
One can thus theoretically derive from the structure underlying these compensatory motor changes a notion of space abstracted from the specific sensory inputs that any given environment's content induces.

This philosophical stance has inspired recent theoretical works on the perception of space, and has in particular been coupled with the SensoriMotor Contingencies Theory (SMCT), a groundbreaking theory of perception that gives prominence to the role of motor information in the emergence of perceptive capabilities \cite{o2001sensorimotor}.
It led to theoretical results regarding the extraction of the dimension of space \cite{Laflaquiere2012}, the characterization of displacements as compensable sensory variations \cite{terekhov2016space}, the grounding of the concept of \emph{point of view} in the motor space \cite{Laflaquiere2013, Laflaquiere2015a}, as well as the characterization of the metric structure of space via sensorimotor invariants \cite{laflaquiere2018discovering}.
These theoretical works suggest that an egocentric concept of space should  emerge first, and that it could be grounded in the motor space as a way to economically capture the sensorimotor invariants that space induces.
Our goal is thus to study how an unsupervised agent can build an internal representation of its sensor's egocentric spatial configuration akin to the $(x,y,z)$ Euclidean description that would otherwise be provided by a hand-designed model. This implies capturing the topology and regular metric structure of the external space in which the sensor moves in a way that does not depend on the content of the environment.
This basic egocentric representation would be a solid foundation for the development of richer spatial knowledge and reasoning (ex: navigation, localization...).

The contribution of this work is to cast the aforementioned theoretical works in an unsupervised (self-supervised) Machine Learning frame. We further develop the formalization of space-induced sensorimotor invariants, and show that a representation capturing the topological and metric structure of space can emerge as a by-product of sensorimotor prediction.
These results shed new light on the fundamental nature of spatial perception, and give some insight on the autonomous grounding of spatial knowledge in a naive agent's sensorimotor experience.

\section{Related work}
\label{sec:Related work}

Only a quick overview of the large literature related to spatial representation learning is given here, leaving aside approaches where the spatial structure of the problem is largely hard-coded \cite{cadena2016past}.

The problem is often conceptualized as the learning of grid or place cells, inspired by neuroscience \cite{best2001spatial}.
Place cells have been built as a way to compress sensory information \cite{arleo2001place}, or to improve sensorimotor and reward predictability \cite{weiller2010involving, stachenfeld2017hippocampus, gustafson2011grid}.
Grid cells have been built as an intermediary representations in recurrent networks trained to predict an agent's position \cite{banino2018vector, cueva2018emergence}.
Both place and grid cells have also been extracted by processing the internal state of a reservoir \cite{antonelo2009towards}.
Representations akin to place cells and displacements have also been built from low-level sensorimotor interaction \cite{kuipers2000spatial}.
Theses works rely on the extraneous definition of ``spatial'' inductive bias or hand-designed loss functions.
\\
In RL, state representation learning is often used to solve spatial tasks (ex: navigation). Some noteworthy works build states based on physical priors \cite{Jonschkowski2015}, system controllability \cite{watter2015embed}, action sequencing \cite{bowling2005action}, or disentanglement of controllable factors in the data \cite{thomas2017independently}.
Many end-to-end approaches are also applied to spatial problems without explicitly building spatial representations \cite{mnih2013playing, kahn2017self}, although auxiliary tasks are sometimes used to induce spatial constraints during training \cite{mirowski2016learning}.
These works once again rely on hand-designed priors to obtain spatial-like representations.
\\
Like in this work, forward sensorimotor predictive models are learned in many methods to compress sensory inputs, improve policy optimization, or derive a curiosity-like reward \cite{ha2018world, eslami2018neural, wayne2018unsupervised, pathak2017curiosity}. Such forward models are also at the core of body schema learning approaches \cite{hoffmann2010body, kwiatkowski2019task}.
Closer to this work, an explicit representation of displacements is built in  \cite{ortiz2018learning} by integrating motor sequences for sensory prediction.
However these works do not study how spatial structure can be captured in such models.
\\
Different flavors of Variational Auto-Encoders have been used to encode ``spatial'' factors of variation in a latent representation \cite{higgins2016beta}; a work that interestingly led to the definition of disentanglement of spatial factors as \emph{invariants} in an agent's experience \cite{higgins2018towards}. These works however ignore the motor component of the problem.
\\
Finally, this work is in line with the theoretical developments of \cite{philipona2003there, Laflaquiere2012, Laflaquiere2013, Laflaquiere2015a, terekhov2016space, laflaquiere2018discovering, marcelmove}, which address the fundamental problem of space perception in the framework of the SMCT, but frame them in an unsupervised machine learning framework.
We show that the structure of space can get naturally captured as a by-product of sensorimotor prediction.

\section{Problem setup}
\label{sec:Problem setup}

Let's consider an agent and an environment immersed in space. The agent has a fixed base, and is equipped with a sensor that it can move to explore its environment. It has only access to raw sensorimotor experiences, and has no a priori knowledge about the world and the external space.
Let $\mathbf{m} \in \mathbb{R}^{N_m}$ be the static configuration of its motors\footnote{If the body is not controlled in position, $\mathbf{m}$ can be a proprioceptive reading of the body configuration.}, referred to as \emph{motor state}.
Let $\mathbf{s} \in \mathbb{R}^{N_s}$ be the reading of its exteroceptive sensor, referred to as \emph{sensory state}.
Let $\epsilon \in \mathbb{R}^{N_\epsilon}$ be the state of the environment defining both its spatial and non-spatial properties.
Finally, let $\mathbf{p} \in \mathbb{R}^{N_p}$ be the external position of the sensor in an egocentric frame of reference centered on the agent's base.
This space of positions is assumed to be a typical Euclidean space with a regular topology and metric.
Our goal is to build, from raw sensorimotor experiences $(\mathbf{m}, \mathbf{s})$, an internal representation $\mathbf{h} \in \mathbb{R}^{N_h}$ which captures the topological and metric structure of $\mathbf{p} \in\mathbb{R}^{N_p}$.
\\
Inspired by Poincar\'e's original insight and borrowing from the formalism of \citep{philipona2003there}, we assume that the agent's sensorimotor experience can be modeled as a continuous mapping parametrized by the state of the environment:
$\mathbf{s}=\phi_{\epsilon}(\mathbf{m})$.
The mapping $\phi$ represents all the constraints that the unknown structure of the world imposes on the agent's experience. In particular, it incorporates the structure of the space in which the agent and the environment are immersed.
It has been shown that this structure manifests itself as invariants in the sensorimotor experience \cite{laflaquiere2018discovering}.
We reformulate here these invariants in a more compact way, taking advantage of the continuity of the sensorimotor mapping $\phi$.
An intuitive description of them is given below, with a more complete mathematical derivation in Appendix \ref{sec:Mathematical formalism}.

\paragraph{Topological invariants:}
The topology (and in particular the dimensionality) of $\mathbb{R}^{N_p}$ is a priori different from the one of $\mathbb{R}^{N_m}$ and $\mathbb{R}^{N_s}$.
Yet, assuming no consistent sensory ambiguity between different sensor positions in the environments the agent explores, the sensory space experienced by the agent in each environment is a manifold, embedded in $\mathbb{R}^{N_s}$, which is homeomorphic to the space $\mathbb{R}^{N_p}$.
Intuitively, this means that small displacements of the sensor are associated with small sensory changes, and vice versa, for any environmental state $\epsilon$.
From a motor perspective, this implies that motor changes associated with small sensory changes correspond to small external displacements:
\begin{equation}
\forall \epsilon, \:
| \phi_{\epsilon}(\mathbf{m}_t) - \phi_{\epsilon}(\mathbf{m}_{t+1}) | \ll \mu
\Leftrightarrow \:
| \mathbf{s}_t - \mathbf{s}_{t+1} | \ll \mu
\Leftrightarrow \:
| \mathbf{p}_{t} - \mathbf{p}_{t+1} | \ll \mu, 
\label{eq:topo}
\end{equation}
where $|.|$ denotes a norm, and $\mu$ is a small value.
The topology of $\mathbb{R}^{N_p}$ is thus accessible via sensorimotor experiences, and constrains how different motor states get mapped to similar sensory states.
In the particular case of a redundant motor system, the multiple $\mathbf{m}$ which lead to the same sensor position $\mathbf{p}$ all generate the same sensory state $\mathbf{s}$ for any environmental state $\epsilon$. The agent has thus access to the fact that the manifold of sensory states, and thus the space of sensor positions, is of lower dimension than the one of its motor space.
Note that these relations are invariant to the environmental state $\epsilon$.
\\
We hypothesize that these \emph{topological invariants} should be accessible to the agent under \textbf{condition I}: when exploring the world, the agent should experience \emph{consistent} sensorimotor transitions $(\mathbf{m}_{t}, \mathbf{s}_{t}) \rightarrow (\mathbf{m}_{t+1}, \mathbf{s}_{t+1})$ such that the state of the environment $\epsilon$ stays unchanged during the transition.

\paragraph{Metric invariants:}
The metric of $\mathbb{R}^{N_p}$ is a priori different from the metric of $\mathbb{R}^{N_m}$ and  $\mathbb{R}^{N_s}$.
Yet, the metric regularity of the external space is accessible in the sensorimotor experience if the environment also undergoes displacements \citep{laflaquiere2018discovering}.
Indeed, let's consider two different sensory states $\mathbf{s}_{t}$ and $\mathbf{s}_{t+1}$ associated with two motor states $\mathbf{m}_{t}$ and $\mathbf{m}_{t+1}$ when the environment is in a first position $\epsilon$. The same sensory states can be re-experienced with two different motor states $\mathbf{m}_{t'}$ and $\mathbf{m}_{t'+1}$ after the environment moved rigidly to a new position $\epsilon'$.
This is the \emph{compensability} property of displacements coined by H.Poincar\'e.
Thus, the consequence of the environment moving rigidly relatively to the agent's base is that equivalent displacements of the sensor in the external Euclidean space $\overrightarrow{\mathbf{p}_{t}\mathbf{p}_{t+1}} = \overrightarrow{\mathbf{p}_{t'}\mathbf{p}_{t'+1}}$ can generate the same sensory change $\mathbf{s}_{t} \rightarrow \mathbf{s}_{t+1}$ for different positions of the environment.
In turn, the different motor changes $\mathbf{m}_{t} \rightarrow \mathbf{m}_{t+1}$ and $\mathbf{m}_{t'} \rightarrow \mathbf{m}_{t'+1}$ generating equivalent sensor displacements are associated with the same sensory changes for different positions of the environment.
It ensues that (see Appendix \ref{sec:Mathematical formalism} for the complete development):
\begin{equation}
\forall \epsilon, \epsilon',
\begin{cases}
| \phi_{\epsilon}(\mathbf{m}_t) - \phi_{\epsilon'}(\mathbf{m}_{t'}) | \ll \mu
\\
| \phi_{\epsilon}(\mathbf{m}_{t+1}) - \phi_{\epsilon'}(\mathbf{m}_{t'+1}) | \ll \mu
\end{cases}
\Leftrightarrow
\;\;\; | (\mathbf{p}_{t+1} - \mathbf{p}_{t}) - (\mathbf{p}_{t'+1} - \mathbf{p}_{t'}) | \ll \mu.
%\begin{cases}
%\big| | \mathbf{p}_{t'} - \mathbf{p}_{t} | - | \mathbf{p}_{t'+1} - \mathbf{p}_{t+1} | \big| \ll \mu \\
%\big| | \mathbf{p}_{t'} - \mathbf{p}_{t'+1} | - | \mathbf{p}_{t} - \mathbf{p}_{t+1} | \big| \ll \mu
%\end{cases} .
\label{eq:metric}
\end{equation}
These relations are once again invariant to the states $\epsilon$ and $\epsilon'$, as long as $\epsilon \rightarrow \epsilon'$ corresponds to a global rigid displacement of the environment.
The metric regularity of $\mathbb{R}^{N_p}$ is thus accessible via sensorimotor experiences, and constrains how different motor changes get mapped to similar sensory changes, for different positions of the environment.
\\
We hypothesize that these \emph{metric invariants} should be accessible to the agent under \textbf{condition II}: the agent should experience \emph{displacements of the environment} $\epsilon \rightarrow \epsilon'$ \emph{in-between} consistent sensorimotor transitions $(\mathbf{m}_{t}, \mathbf{s}_{t}) \rightarrow (\mathbf{m}_{t+1}, \mathbf{s}_{t+1})$ and $(\mathbf{m}_{t'}, \mathbf{s}_{t'}) \rightarrow (\mathbf{m}_{t'+1}, \mathbf{s}_{t'+1})$.

Space thus induces an underlying structure in the way motor states map to sensory states.
Interestingly, the sensory component of this structure varies for different environments, but not the motor one: a single $\mathbf{m}$ is always associated with the same egocentric position $\mathbf{p}$, regardless of the environmental state $\epsilon$, while the associated $\mathbf{s}$ changes with $\epsilon$.
A stable representation of $\mathbf{p}$ can then be grounded in the motor space (as already argued in \cite{Laflaquiere2015a}), and shaped by sensory experiences which reveal this structure.
%The work of \cite{laflaquiere2018discovering} showed that it is indeed theoretically possible to build such a representation $\mathbf{h}$ by having i) motor states $\mathbf{m}$ associated with similar sensory states $\mathbf{s}$, for any $\epsilon$, represented by similar representations $\mathbf{h}$, and ii) motor transitions $\mathbf{m} \rightarrow \mathbf{m}'$ associated with similar sensory transitions $\mathbf{s}\rightarrow\mathbf{s}'$, for different $\epsilon$, represented by similar representation transitions $\mathbf{h}\rightarrow\mathbf{h}'$.
\\
In this work, we hypothesize that capturing space-induced invariants is beneficial for sensorimotor prediction, as they can act as acquired priors over sensorimotor transitions no yet experienced.
For instance, imagine two motor states $\mathbf{m}_a$ and $\mathbf{m}_b$ have always been associated with identical sensory states in the past. Then encoding them with the same representation $\mathbf{h}_{a} = \mathbf{h}_{b}$ can later help the agent extrapolate that if $\mathbf{m}_a$ is associated with a previously unseen sensory state $\mathbf{s}_a$, then $\mathbf{m}_b$ will also be.
Therefore, we propose to train a neural network to perform sensorimotor prediction, and to analyze how it learns to encode motor states depending on the type of exploration that generates the sensorimotor data. We expect this learned representation to capture the topology of $\mathbf{p} \in \mathbb{R}^{N_p}$ when condition I is fulfilled, and to capture its metric regularity when condition II is fulfilled, without extraneous constraint nor supervision.

\section{Experiments}
\label{sec:Experiments}

\paragraph{Sensorimotor predictive network:}
\label{sec:Sensorimotor predictive network}

\begin{figure*}[t]
\centering
\includegraphics[width=0.99\linewidth]{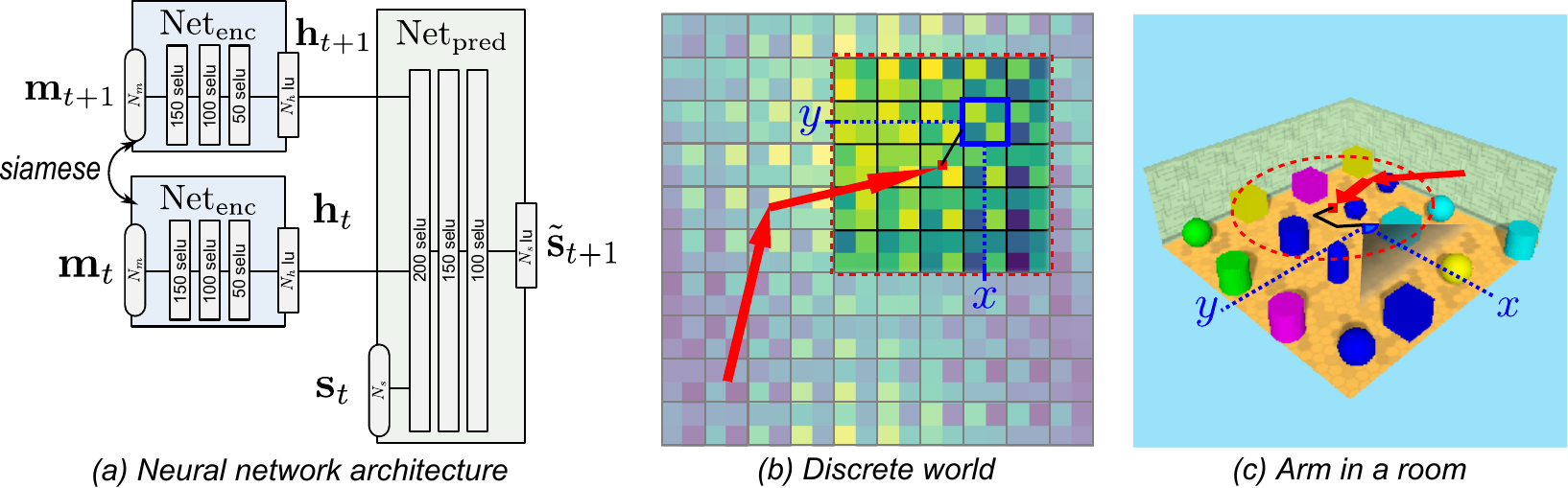}
\caption{(a) The neural network architecture featuring two siamese instances of the $\text{Net}_\text{enc}$ module, and a $\text{Net}_\text{pred}$ module.
(b-c) Illustrations of the Discrete world and Arm in a room simulations. The fixed base of the agent is represented by a red square. The working space reachable by the sensor is framed in dotted red. The sensor and its current egocentric position $\mathbf{p}=[x,y]$ in this frame are displayed in blue.
The environment can translate relatively to the agent's base, which is equivalent to an (opposite) displacement of the agent's base itself, as illustrated by red arrows. (Best seen in color)}
\label{fig:setups}
\end{figure*}

We propose a simple neural network architecture to perform sensorimotor prediction. The network's objective is to predict the sensory outcome $\mathbf{s}_{t+1}$ of a future motor state $\mathbf{m}_{t+1}$, given a current motor state $\mathbf{m}_t$ and sensory state $\mathbf{s}_t$. Additionally, we want both motor states $\mathbf{m}_t$ and $\mathbf{m}_{t+1}$ to be encoded in the same representational space.
As illustrated in Fig.~\ref{fig:setups}, the network is thus made of two types of modules:
i) $\text{Net}_\text{enc}$, a Multi-Layer Perceptron (MLP) taking a motor state $\mathbf{m}_t$ as input, and outputting a motor representation $\mathbf{h}_t$ of dimension $N_h$, and
ii) $\text{Net}_\text{pred}$, a MLP taking as input the concatenation of a current representation $\mathbf{h}_t$, a future representation $\mathbf{h}_{t+1}$, and a current sensory state $\mathbf{s}_t$, and outputting a prediction for the future sensory state $\mathbf{\tilde{s}}_{t+1}$.
The overall network connects a predictive module $\text{Net}_\text{pred}$ to two siamese copies of a $\text{Net}_\text{enc}$ module, ensuring that both $\mathbf{m}_t$ and $\mathbf{m}_{t+1}$ are encoded the same way.
The loss function to minimize is the Mean Squared Error (MSE) between the sensory prediction and the ground truth:
\begin{equation}
\text{Loss} = \frac{1}{K} \sum_{k=1}^K |\mathbf{\tilde{s}}_{t+1}^{(k)} - \mathbf{s}_{t+1}^{(k)}|^2,
\end{equation}
where $K$ is the number of sensorimotor transitions collected by the agent.
No extra component is added to the loss regarding the structure of the representation $\mathbf{h}$.
Unless stated otherwise, the dimension $N_h$ is arbitrarily set to $3$ for the sake of visualization. A more thorough description of the network and training procedure is available in Appendix \ref{sec:Neural network and training}.

\paragraph{Analysis of the motor representation:}
\label{sec:Analysis of the motor representation}

We use two measures $D_{topo}$ and $D_{metric}$ to assess how much the structure of the representation $\mathbf{h}$ built by the network differs from the one of the sensor position $\mathbf{p}$.
The first corresponds to an estimation of the topological dissimilarity between a set in $\mathbb{R}^{N_p}$ and the corresponding set in $\mathbb{R}^{N_h}$: 
\begin{equation}
D_{topo} =
\frac{1}{N^2}
\sum_{i,j}^N 
\frac{\big| \mathbf{h}_{i} - \mathbf{h}_{j} \big|}
{\max_{kl} \Big( \big| \mathbf{h}_{k} - \mathbf{h}_{l} \big| \Big)}
. \exp\left( \frac{ - \alpha . \big| \mathbf{p}_{i} - \mathbf{p}_{j} \big| }
{ \max_{kl} \Big( \big| \mathbf{p}_{k} - \mathbf{p}_{l} \big| \Big) } \right),
\label{eq:d_topo}
\end{equation}
where $|.|$ denotes the Euclidean norm, $N$ is the number of samples in each set, and $\alpha$ is arbitrarily set to $50$. This measure is large when close sensor positions $\mathbf{p}$ are encoded by distant motor representations $\mathbf{h}$, and small otherwise.
\\
The second measure corresponds to an estimation of the metric dissimilarity between those same two sets.
At this point, it is important to notice that the metric invariants described in Sec.~\ref{sec:Problem setup} only imply relative distance constraints (see also \cite{laflaquiere2018discovering}). Consequently, any representation $\mathbf{h}$ related to $\mathbf{p}$ via an affine transformation would respect these constraints\footnote{An affine transformation preserves topology and distance ratios.}.
In order to properly assess if the two sets share the same metric regularity, we first perform a linear regression of $\mathbf{p}$ on $\mathbf{h}$ to cancel out the potential affine transformation between the two. We denote $\mathbf{h}^{(p)} = A.\mathbf{h} + b$ the resulting projection of $\mathbf{h}$ in $\mathbb{R}^{N_p}$, where $A$ and $b$ are the optimal parameters of the linear regression.
The second dissimilarity $D_{metric}$ is then defined as:
\begin{equation}
D_{metric} =
\frac{1}{N^2}
\sum_{i,j}^N 
\frac{\Big| \: \big| \mathbf{h}_{i}^{(p)} - \mathbf{h}_{j}^{(p)} \big| - \big| \mathbf{p}_{i} - \mathbf{p}_{j} \big|
\: \Big|}
{\max_{kl} \Big( \big| \mathbf{p}_{k} - \mathbf{p}_{l} \big| \Big)}.
\label{eq:d_metric}
\end{equation}
This measure is large when the distance between two sensor positions $\mathbf{p}$ differs from the distance between the two corresponding motor representations $\mathbf{h}$ (after affine projection), and is small otherwise. It is equal to zero when there exists a perfect affine mapping between $\mathbf{h}$ and $\mathbf{p}$; in which case the two sets have equivalent metric regularities.
\\
Note that in \eqref{eq:d_topo} and \eqref{eq:d_metric}, distances are normalized by the largest distance in the corresponding space in order to avoid undesired scaling effects.
In the following, the dissimilarities are computed on sets of $\mathbf{p}$ and corresponding $\mathbf{h}$ generated by sampling the motor space in a fixed and regular fashion (see Fig.~\ref{fig:projection}). This ensures a rigorous comparison of their values between epochs and between runs.

\paragraph{Types of exploration:}
\label{sec:Types of exploration}

Three types of exploration of the environment are considered to test the hypotheses laid out in Sec.~\ref{sec:Problem setup}. They correspond to different ways to generate the sensorimotor transitions $(\mathbf{m}_t,\mathbf{s}_t) \rightarrow(\mathbf{m}_{t+1},\mathbf{s}_{t+1})$ fed to the network during training:

\textit{Inconsistent transitions in a moving environment}:
The motor space is randomly sampled, and the environment randomly moves between $t$ and $t+1$.
Both conditions I and II are broken, as the agent explores a constantly moving environment, such that its sensorimotor transitions have no spatiotemporal consistency\footnote{It is akin to the kind of data a passive and non-situated agent receives in typical machine learning settings.}.
We refer to this type of exploration as \textbf{MEM} (Motor-Environment-Motor), in agreement with the order of changes for each transition.

\textit{Consistent transitions in a static environment}:
The motor space is sampled randomly, and the environment stays static.
Condition I is fulfilled, as the agent experiences spatiotemporally consistent transitions in a static environment, but not condition II, as the environment does not move between transitions.
We refer to this type of exploration as \textbf{MM} (Motor-Motor).

\textit{Consistent transitions in a moving environment}:
The motor space is randomly sampled, and the environment randomly moves after each transition (after $t+1$).
Both conditions I and II are fulfilled, as the agent experiences spatiotemporally consistent transitions, and the environment moves between transitions.
We refer to this type of exploration as \textbf{MME} (Motor-Motor-Environment).

Additional details on the sampling procedure are available in Appendix \ref{sec:Neural network and training}.
According to Sec.~\ref{sec:Problem setup}, we expect the sensorimotor data to contain no spatial invariants in the MEM case, topological invariants in the MM case, and both topological and metric invariants in the MME case.

\paragraph{Agent-Environment setups:}
\label{sec:Agent-Environment setups}

We simulate two different agent-environment setups: \textit{Discrete world} and \textit{Arm in a room}.
The first one is a minimalist artificial setup designed to test our hypotheses in optimal conditions.
It corresponds to a 2D grid-world environment that the agent can explore by generating 3D motor states $\mathbf{m}$ which map, in a non-linear and redundant way, to 2D positions $\mathbf{p}$ of the sensor in the grid (see Fig.~\ref{fig:setups}(b)).
For each position the agent receives a 4D sensory input $\mathbf{s}$ that is designed to vary smoothly over the grid and to have no sensory ambiguity between different positions.
The whole grid can translate with respect to the agent's base, changing its state $\epsilon$, and acts as a torus to avoid any border effect.
\\
The second one is a more complex and realistic setup in which a three-segment arm equipped with a camera explores a 3D room filled with random objects (see Fig.~\ref{fig:setups}(c)).
The agent can change its camera position $\mathbf{p}$ in a horizontal plane (with a fixed orientation) by generating 3D motor states $\mathbf{m}$, and receives sensory inputs $\mathbf{s}$ of size $768$ ($16\times16$ RGB pixels).
The whole room can translate in 2D with respect to the agent's base, changing its state $\epsilon$, and the arm cannot move outside of the room.
\\
A more complete description of the simulations is available in Appendix \ref{sec:Simulations}.

\begin{figure*}[t!]
\centering
\includegraphics[width=1\linewidth]{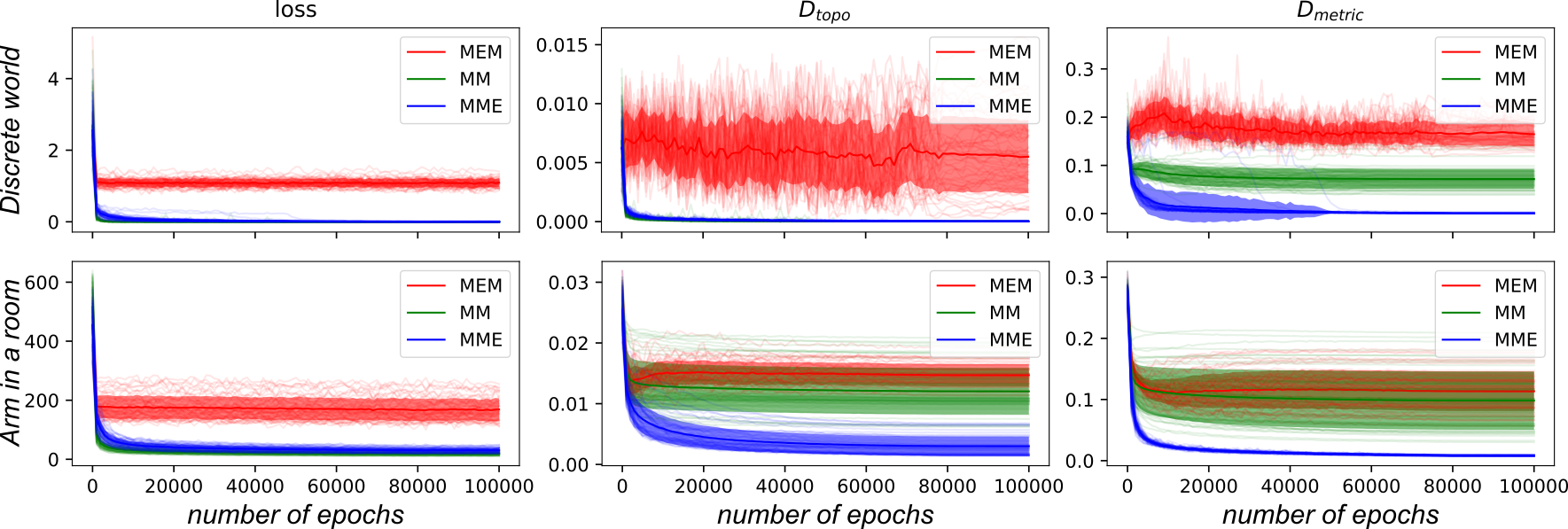}
\caption{Evolution of the loss and the dissimilarity measures $D_{topo}$ and $D_{metric}$ during training for both setups, and for the three types of exploration. The displayed means and standard deviations are computed over 50 independent runs. (Best seen in color)}
\label{fig:results}
\end{figure*}

%\textit{Arm in a room}:
%A more complex setup of an arm exploring a room full of objects, implemented using the Flatland simulator \citep{caselles2018flatland}. As illustrated in Fig.~\ref{fig:setups}, the environment is a 2D square room of width 12 units with walls, filled with simple geometric objects (squares, circles, triangles) with random properties (number, size, position, color).
%The agent is a three-segment arm equipped with an array of $10$ distance sensors at its end, generating a sensory state $\mathbf{s}$ of dimension $N_s=10$. The three arm segments are each of length 1 unit, and their respective relative orientation are controlled in $[-\pi,\pi]$ radians by three independent motors, giving a motor state $\mathbf{m}$ of dimension $N_m=3$. The effective working space in which the sensor can change its egocentric position $\mathbf{p} = [x, y]$ of dimension $N_p=2$ is thus a disk of radius $3$ units. The sensor's orientation is kept fixed, such that the sensor only translates in the working space.
%Finally, the environment can translate relatively to the arm's base with a maximal horizontal and vertical range of $[-3, 3]$, where a translation of $[0, 0]$ corresponds to the room being centered on the agent's base. The agent cannot move its sensor beyond the walls.

\section{Results}
\label{sec:Results}

\begin{figure*}[t!]
\centering
\includegraphics[width=1\linewidth]{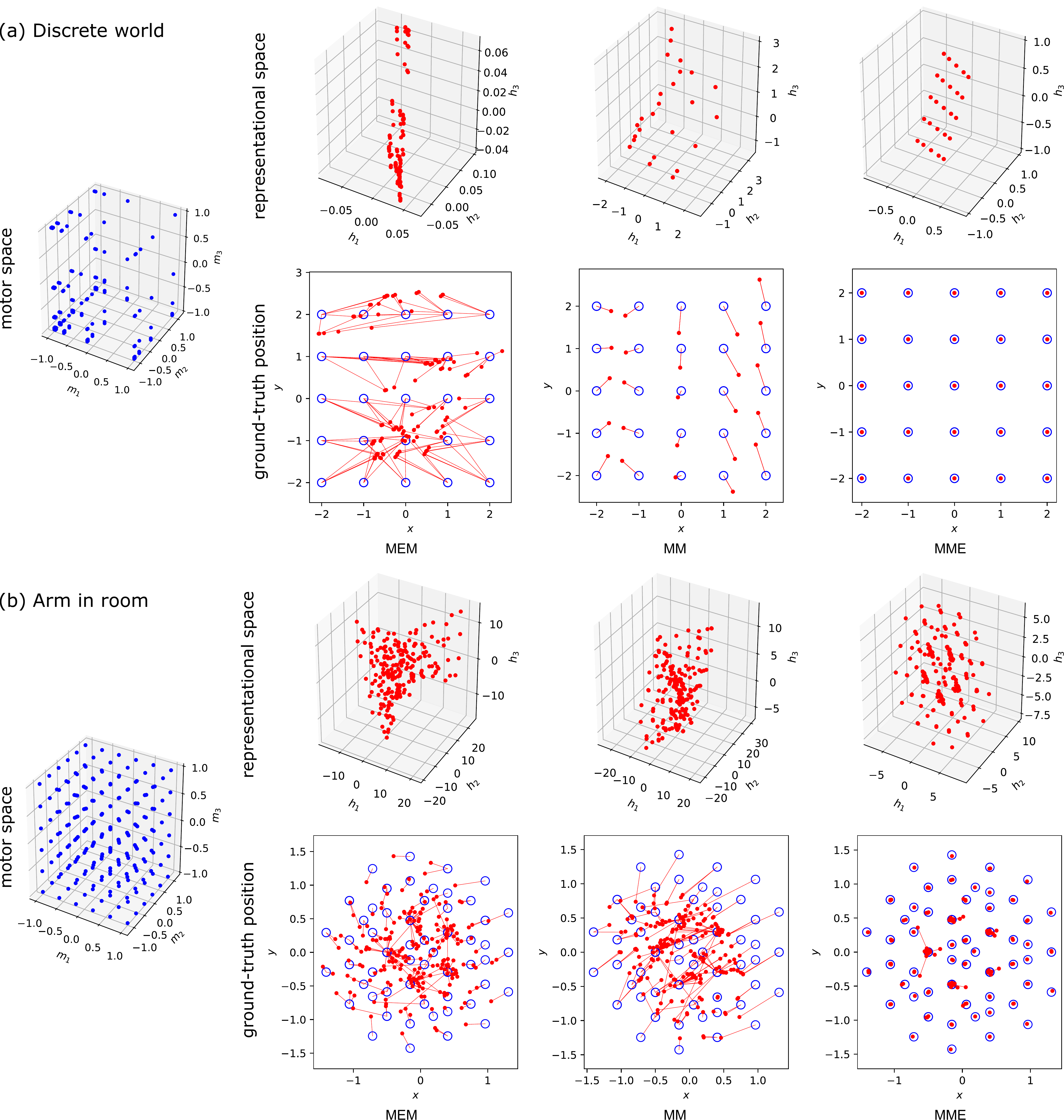}
\caption{Visualization of the normalized regular motor sampling $\mathbf{m}$ (blue dots), its representation $\mathbf{h}$ in the representational space (red dots), and the corresponding ground truth position $\mathbf{p}$ (blue circles) for the three types of exploration, and for both the Discrete world (a) and Arm in a room (b) setups. The linear regression $\mathbf{h}^{(p)}$ of the representations are also displayed in the space of positions. Lines have been added to visualize the distances between each $\mathbf{h}^{(p)}$ and its ground truth counterpart $\mathbf{p}$.
In (a), the regular (but non-linear) 3D motor sampling generates a regular 2D grid of positions, where each position corresponds to 5 redundant motor states. In (b), the regular 3D motor sampling generates a star-shaped set of positions, where some positions, like the inner corners of the star, correspond to multiple motor states. (Best seen in color)}
\label{fig:projection}
\end{figure*}

We evaluate the three types of exploration on the two experimental setups.
Each simulation is run 50 times, with all random parameters drawn independently on each trial.
During training, the measures $D_{topo}$ and $D_{metric}$ are evaluated on a fixed regular sampling of the motor space. Their evolution, as well as the evolution of the loss, are displayed in Fig.~\ref{fig:results}.
Additionally, Fig.~\ref{fig:projection} shows the final representations $\mathbf{h}$ of the regular motor sampling, for one randomly selected trial of each simulation. The corresponding positions $\mathbf{p}$ and the projection $\mathbf{h}^{(p)}$ are displayed in the same space in order to visualize how much their metric structures differ.

\paragraph{Discrete world results:}
\label{sec:Discrete world results}

The results clearly show an impact of the type of exploration on the motor encoding built by the network.
First of all, as expected, the loss is high in the MEM case because the constant movements of the environment prevent any accurate sensorimotor prediction.
On the contrary, it is low in the MM and MME cases, as the consistency of transitions enables accurate sensorimotor prediction (upper-bounded by the expressivity power of the $\text{Net}_{\text{pred}}$ module).
\\
More interestingly, the topological dissimilarity $D_{topo}$ has a significantly smaller value and variance in the MM and MME cases than in the MEM case.
This seems to indicate that the MEM exploration leads to arbitrary representations, while the topologies of $\mathbf{h}$ and $\mathbf{p}$ are more similar when the agent can experience consistent sensorimotor transitions $(\mathbf{m}_t,\mathbf{s}_t) \rightarrow(\mathbf{m}_{t+1},\mathbf{s}_{t+1})$ during which the environment does not move.
\\
Similarly, the metric dissimilarity $D_{metric}$ is high in the MEM case, average in the MM case, and low in the MME case.
This intermediate value in the MM case is due to the fact that capturing the topology of $\mathbf{p}$ also indirectly reduces the metric dissimilarity.
However, the very low value in the MME case seems to indicate that the metric of $\mathbf{h}$ displays a regularity which is similar to the one of $\mathbf{p}$. This regularity is thus captured only when the agent can experience movements of the environment between consistent sensorimotor transitions.
\\
This analysis is confirmed in Fig.~\ref{fig:projection} where $\mathbf{h}$ and its affine projection $\mathbf{h}^{(p)}$ in $\mathbb{R}^{N_p}$ display an arbitrary structure in the MEM case, a structure topologically equivalent to the one of $\mathbf{p}$ in the MM case, and a structure topologically and metrically equivalent to the one of $\mathbf{p}$ in the MME case.

%\emph{MEM exploration:}
%The loss and measures $Q_{topo}$ and $Q_{metric}$ stay at relatively high values throughout the training.
%Such a high loss indicates that the agent is unable to learn a good sensorimotor predictive mapping, which is expected as the environment moves during each sensorimotor transition. The current pair $(\mathbf{m}_t, \mathbf{s}_t)$ is therefore not informative to predict $\mathbf{s}_{t+1}$.
%Moreover, the high final $Q_{topo}$ and $Q_{metric}$ values indicate that the structure of the representation $\mathbf{h}$ significantly differs from the structure of the sensor position $\mathbf{p}$.
%This is confirmed in Fig.~\ref{fig:projection} where we can see that the structure of $\mathbf{h}$ does not match the one of $\mathbf{p}$, metric-wise or even topology-wise. In particular, redundant motor states $\mathbf{m}$, corresponding to the same sensor position $\mathbf{p}$, are represented by different $\mathbf{h}$.

\paragraph{Arm in a room results:}
\label{sec:Arm in a room results}

The results in this more complex and realistic setup are globally similar to the previous ones, which shows a consistency in the phenomenons we just described.
However, two important differences can be pointed out. 
\\
The first is that, in the MEM case, $D_{topo}$ and $D_{metric}$ seem to decrease more than what would be expected for an arbitrarily random representation $\mathbf{h}$.
We argue that this phenomenon is due to a ``border effect'' induced by the walls of the room.
Indeed, as the sensor's and environment's displacements are limited, each motor state $\mathbf{m}$ is statistically associated with a different distribution of sensory states over the whole course of the exploration.
For instance, a motor state corresponding to the arm extended to the right will never experience the sensory states corresponding to the sensor being at the far-left of the room, and inversely. As the sensory distribution associated with $\mathbf{m}$ varies smoothly with $\mathbf{p}$, the agent can indirectly infer the topology of $\mathbb{R}^{N_p}$ from it.
We can indeed see in Fig.~\ref{fig:projection} that $\mathbf{h}$ (or even better $\mathbf{h}^{p}$) tends to capture the topology of $\mathbf{p}$ in the MEM case, although with less accuracy than in the MM and MEM cases.
Note that this was not the case in the previous torus-like grid world as any motor state could be associated with any square of the grid over the course of the exploration.
\\
The second difference is that $D_{topo}$ is higher in the MM case than in the previous setup.
After an empirical visualization of the environments and associated learned representations $\mathbf{h}$, we argue that this is due to potential sensory ambiguity.
Indeed, the random room the agent explores can present ambiguities, such that very different positions of the sensor can be associated with very similar sensory inputs. When this happens, the representation $\mathbf{h}$ built by the network can arbitrarily encode the same way different motor states associated with these different positions. This leads to a representation manifold that is non-trivially twisted in $\mathbb{R}^{N_h}$ (see Fig.~\ref{fig:projection}(b)), and degrades the measure $D_{topo}$.
Note that this sensory ambiguity is not an issue in the MME case anymore, as the movements of the environment help disambiguate these different motor states.

The same experiments have been run with a representational space of dimension $N_h = 25$ and with more complex agent morphologies, and led to qualitatively similar results. This seems to indicate that the capture of the topological and metric invariants is insensitive to the dimension of $\mathbf{h}$ and the complexity of the forward mapping.
A more detailed analysis of the all these different simulations results is available in Appendix \ref{sec:Detailed results analysis} and \ref{sec:Additional experiments}.

%\subsection{Additional experiment}
%\label{sec:Additional experiment}
%An additional experiment is performed in order to evaluate the robustness of these result with respect to the complexity of the sensorimotor mapping and the dimension of the representational space. The arm is now equipped with a RGB camera of resolution $16$ pixels, and $N_h$ is set to $25$ instead of $3$. The results, displayed in black in Fig.~\ref{fig:results}, are qualitatively identical to the initial setup. This seems to indicates that the emergence of the spatial-like structure in the representation $\mathbf{h}$ is insensitive to the complexity of the sensory input, and to the dimensionality of the representational space (see Appendix \ref{sec:Additional experiment app}).

\section{Conclusion}
\label{sec:Conclusion}

We addressed the problem of the unsupervised grounding of the concept of space in a naive agent's sensorimotor experience.
Inspired by previous philosophical and theoretical work, we argue that such a notion should first emerge as a basic representation of the egocentric position of a sensor moving in space.
We showed that the structure of the Euclidean space, in which the agent is immersed alongside its environment, induces sensorimotor invariants.
They constrain the way motor states gets mapped to sensory inputs, independently from the actual content of the environment, and carry information about the topology and metric of the external space.
This structure can potentially be extracted from the sensorimotor flow to build an internal representation of the sensor egocentric position, grounded in the motor space, and abstracted from the content of the environment and the specific sensory states it induces.
We hypothesized that capturing space-induced invariants is beneficial for sensorimotor prediction.
As a consequence, topological and metric invariants should naturally be captured by a network learning to perform sensorimotor prediction.
We proposed such a network architecture, and designed different types of exploration of the environment such that topological and metric invariants were present or not in the resulting sensorimotor transitions fed to the network.
We tested two different simulated agent-environment setups, and showed that when spatial invariants are present in the sensorimotor data, they get naturally captured in the internal motor representation built by the agent.
So, when the agent can experience consistent sensorimotor transitions during which the environment does not change, the internal motor representation captures the topology of the external space in which its sensor is moving.
Even more interestingly, when the agent can also experience displacements of the environment between its consistent sensorimotor transitions, the internal motor representation captures the metric regularity of this external space.
These results thus suggest that the concept of an external Euclidean space, although still in its most basic form here, could emerge in a situated agent as a by-product of learning to predict its sensorimotor experience.
\\
We hope this work can be a stepping stone for further extensions of the approach and the unsupervised acquisition of richer spatial knowledge.
A first obvious step will be to extend the exploration to 3 translations and 3 rotations of the sensor in space.
A second very important step will be to derive, from the current basic egocentric spatial representation, an allocentric representation in which the spatial configurations of external objects could also be characterized; a problem for which H.Poincar\'e also had some interesting intuitions.

\bibliographystyle{plainnat}
\bibliography{NIPS2019_biblio}

\clearpage

\appendix

\section{Mathematical formalism}
\label{sec:Mathematical formalism}

\subsection{Sensorimotor interaction}

\begin{figure}[h]
\centering
\includegraphics[width=0.9\linewidth]{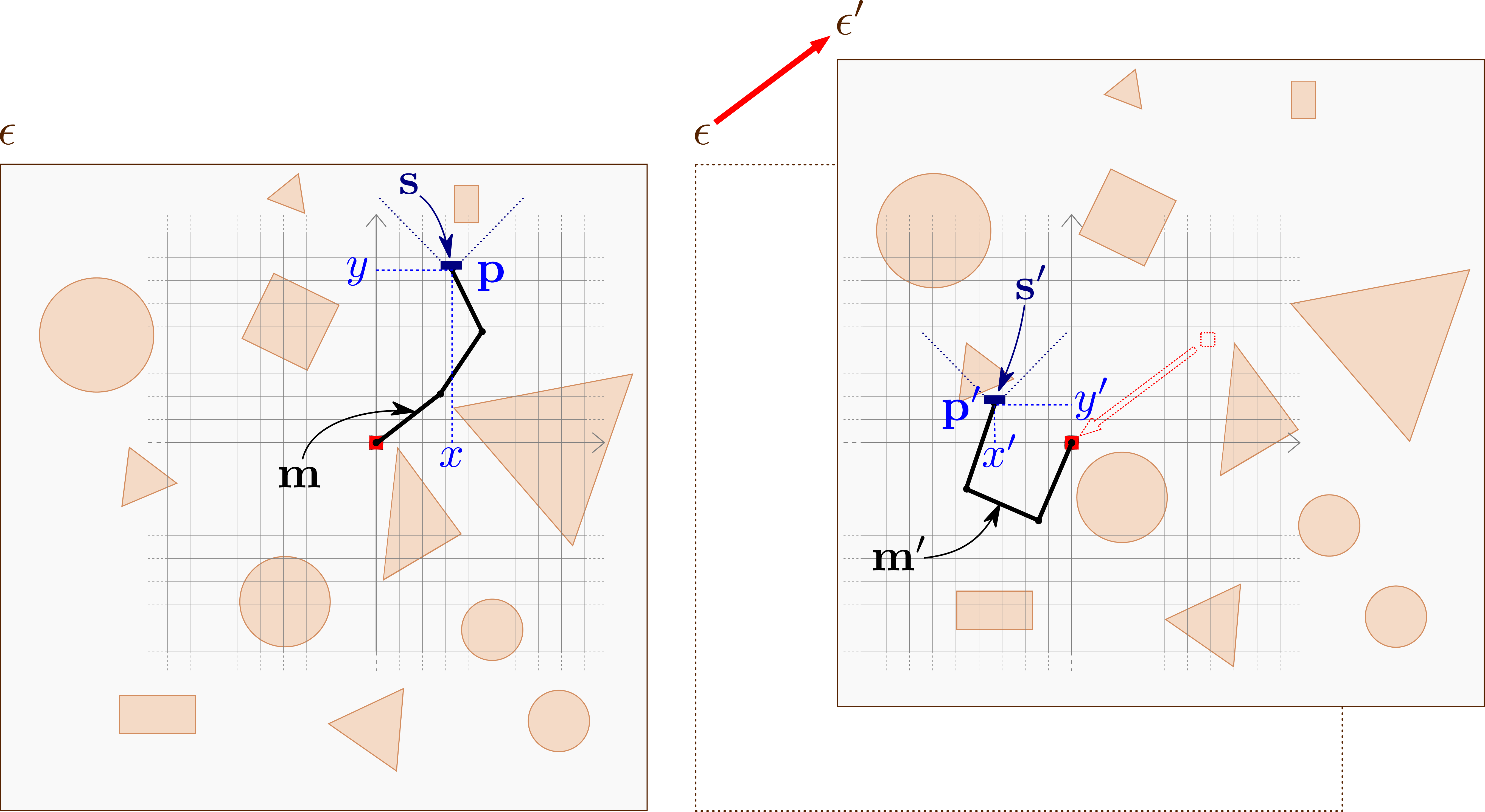}
\caption{Diagram of a three-segment agent (black) with a fixed base (red square) exploring an environment (brown) that can move.
The motor state $\mathbf{m}$ defines the configuration of the arm.
It is associated with a position $\mathbf{p} = [x, y]$ of its end-effector sensor (blue), relative to its base.
This egocentric frame of reference is represented as a grid centered on the agent's basis.
The sensor generates a sensory state $\mathbf{s}$.
The agent can move its sensor in the environment, but the environment can also move (red arrow) relatively to the agent's basis, changing its state $\epsilon$.
Note that such a rigid displacement of the whole environment is equivalent to an opposite displacement of the agent's basis in a static environment (red dotted arrow).
(Best seen in color)}
\label{fig:diagram_agent}
\end{figure}

We consider the sensorimotor interaction between an agent and its environment immersed in an Euclidean space.
The agent has a base with a fixed position in space, and is equipped with an exteroceptive sensor that it can move in space thanks to its motors.
These motors are supposed to be controlled in position, which means that each motor state corresponds to a fixed posture of the agent, and thus to a fixed position of the sensor in space.
The environment itself can move in a rigid fashion, independently from the agent (see Fig.~\ref{fig:diagram_agent}).
\\
In this work, we show how performing sensorimotor prediction naturally leads such an agent to build an internal representation of the egocentric position of its sensor which captures both the topology and metric regularity of the external space.

Note that, from a sensorimotor perspective, a displacement of the environment relative to the (fixed) agent's base is equivalent to a movement of the agent's base relative to the (fixed) environment.
This equivalence is behind the concept of \emph{compensability} introduced by Poincar\'e to characterize spatial interactions \cite{Poincare1895}.
Considering that the agent moves its base relative to a static environment sounds like a more natural description, as we rarely experience rigid displacements of the whole environment around us.
However we favor the first description in this work, as a way to follow Poincar\'e's original insight and to better emphasize the interaction between motor changes and environmental changes in the mathematical formalism.

\subsection{Variables and mappings}

\begin{figure}[t!]
\centering
\includegraphics[width=0.75\linewidth]{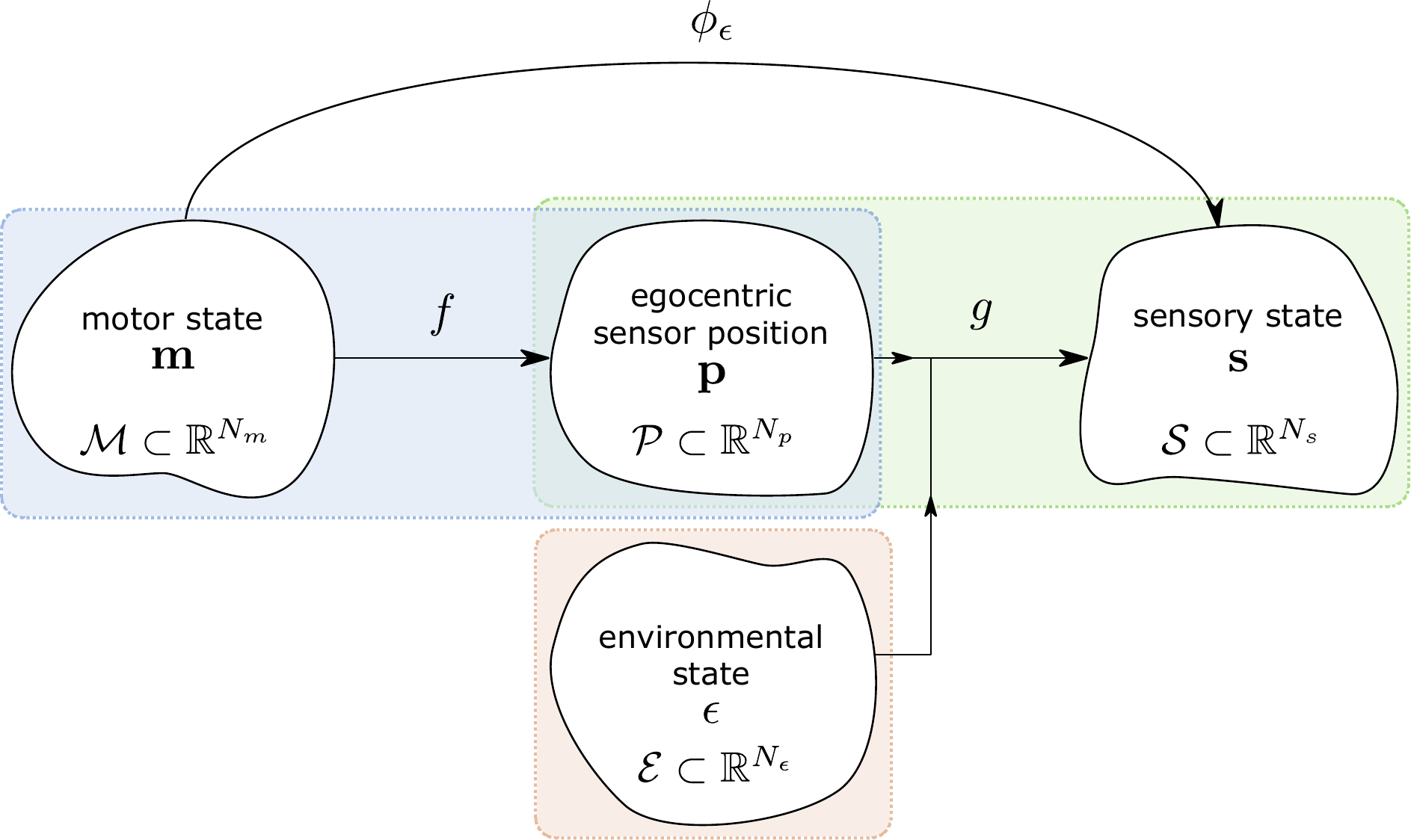}
\caption{Illustration of the variables and mappings involved in the formalism. (Best seen in color)}
\label{fig:mapping_functions}
\end{figure}

We denote $\mathbf{m}$ the agent's motor state, which corresponds to a static posture of its body, and $\mathcal{M}$ the set of all $\mathbf{m}$.
We denote $\mathbf{p}$ the egocentric position of the sensor relative to the agent's base, and $\mathcal{P}$ the set of all $\mathbf{p}$.
We denote $\epsilon$ the state of the environment, which describes both its spatial and non-spatial properties, and $\mathcal{E}$ the set of all $\epsilon$.
Finally, we denote $\mathbf{s}$ the agent's sensory state, which corresponds to an instantaneous reading of the sensor's output without transient phase, and $\mathcal{S}$ the set of all $\mathbf{s}$.
\\
The different relations between these variables are illustrated in Fig.\ref{fig:mapping_functions}.
Each motor state $\mathbf{m}$ is associated with a sensor position $\mathbf{p}$ via a forward mapping denoted $f$:
\begin{align}
\begin{split}
f: \: \mathcal{M} &\to \mathcal{P}\\
   \: \mathbf{m} &\mapsto f(\mathbf{m}) = \mathbf{p}.
\label{eq:f}
\end{split}
\end{align}
Similarly, each pair $(\mathbf{p}, \epsilon)$ of sensor position and environmental state is associated with a sensory state $\mathbf{s}$ via a sensory mapping denoted $g$:
\begin{align}
\begin{split}
g: \: \mathcal{P} &\times\mathcal{E} \to \mathcal{S}\\
   \: \mathbf{p}  &, \epsilon \mapsto g(\mathbf{p}, \epsilon) = \mathbf{s}.
\label{eq:g}
\end{split}
\end{align}
Due to the inherent properties of space, the sensory mapping $g$ has the following invariance property:
\begin{equation}
\forall \delta, \mathbf{p}, \epsilon, \:
g(\mathbf{p}, \epsilon + \delta) = g(\mathbf{p} - \delta, \epsilon),
\label{eq:delta}
\end{equation}
where $\delta$ represents a rigid displacement of either the whole environment or of the sensor.
In other words, from a sensory perspective, a displacement of the environment is equivalent to an opposite displacement of the sensor.

In this work, we consider that the agent is naive. It only has direct access to the motor states it produces and to the related sensory states it receives: $(\mathbf{m}, \mathbf{s})$.
It does not have a priori information about $f$, $g$, or $\epsilon$.
From a sensorimotor perspective, the overall composite mapping $g \circ f$ can thus be re-expressed as a sensorimotor mapping $\phi_\epsilon$:
\begin{equation}
\mathbf{s} = g\big(f(\mathbf{m}), \epsilon\big) = \phi_\epsilon(\mathbf{m}),
\label{eq:phi}
\end{equation}
where $\epsilon$ is treated as a parameter of the unknown mapping $\phi$ to emphasize that the agent has no information about both $\phi$ and $\epsilon$ (see Fig.~\ref{fig:mapping_functions}).

\subsection{Hypotheses}

Taking inspiration from the differential geometry-based formalism introduced in \cite{philipona2003there}, we assume that $\mathcal{M} \subset \mathbb{R}^{N_m}$, $\mathcal{S} \subset \mathbb{R}^{N_s}$, $\mathcal{P} \subset \mathbb{R}^{N_p}$, and $\mathcal{E} \subset \mathbb{R}^{N_\epsilon}$ are manifolds embedded in their respective finite vector spaces. The variables $\mathbf{m}$, $\mathbf{s}$, $\mathbf{p}$, $\epsilon$ can thus be expressed as real-valued vectors. Additionally, we posit that $\mathcal{M}$ and $\mathcal{P}$ are convex bounded Euclidean subspaces of their respective dimensions, to take into account that the agent has a limited body and thus a limited working space in which it can move its sensor.
\\
Moreover, we make the strong assumption that the mappings $f$, $g$ (and consequently $\phi$) are continuous, which means that an infinitesimally small change in their input leads to an infinitesimally small change in their output.
Without any loss of generality, we assume that $f$ is surjective, in order to accommodate for redundant motor systems for which multiple motor states $\mathbf{m}$ can be associated with the same sensor position $\mathbf{p}$.
Finally, for convenience, let $g_\epsilon$ denote the sensory mapping associated with a fixed environmental state $\epsilon$:
\begin{align}
\begin{split}
\forall \epsilon, g_\epsilon: \: \mathcal{P} &\to \mathcal{S}\\
            \: \mathbf{p} &\mapsto g_\epsilon(\mathbf{p}) = g(\mathbf{p}, \epsilon) = \mathbf{s}.
\label{eq:ge}            
\end{split}
\end{align}
In any environment, we assume that there is no sensory ambiguity between two different sensor positions.
Formally, the mapping $g_\epsilon$ is thus assumed to be bijective\footnote{Note that this assumption would not hold for $g$, as suggested by Eq.~\eqref{eq:delta}.} for all $\epsilon$:
\begin{equation}
\forall \epsilon, \mathbf{s}, \exists! \mathbf{p} \;\text{such that:}\; g_\epsilon(\mathbf{p}) = \mathbf{s},
\label{eq:bijection}
\end{equation}
making it a homeomorphism between $\mathcal{P}$ and $\mathcal{S}$, as it is also continuous.
This means that, for a given state $\epsilon$ of the environment, each position of the sensor $\mathbf{p}$ is associated with a unique sensory state $\mathbf{s}$, and vice versa. This hypothesis is expected to hold if the sensor is rich enough and if the environment does not present symmetries.
This apparently strong constraint of the model turns out to be relatively weak in practice. Indeed, thanks to the statistical machine learning approach used in this work, this non-ambiguity assumption has to hold statistically over all environmental states. In other words, it is sufficient that two sensor positions lead to different sensory states for at least \textit{some} environmental states $\epsilon$, but not necessarily all of them.
In strongly unfavorable scenarios, one could also consider extending the model by integrating sensorimotor experiences in time to avoid ambiguities.

\subsection{Sensorimotor invariants:}

The manifold of sensor positions $\mathcal{P}$ has a priori a different topology and metric than the motor manifold $\mathcal{M}$.
For instance, in the case of a redundant motor system, a single sensor position $\mathbf{p} \in \mathcal{P}$ is associated with a subset of dimension greater than 0 in $\mathcal{M}$. As a consequence, the dimension of $\mathcal{P}$ is lower than the one of $\mathcal{M}$ and their topologies differ.
Similarly, the same change (same direction and amplitude) of sensor position in the external space is a priori associated with different motor changes (different directions and amplitudes), depending on the starting sensor position (and vice versa). The metrics of $\mathcal{P}$ and $\mathcal{M}$ thus differ in a non-linear way.
\\
Yet, it has been shown in \cite{laflaquiere2018discovering} that the topology and metric of $\mathcal{P}$ induce invariants in the sensorimotor experiences of the agent.
Here, we rigorously reformulate these invariants, and extend them to small neighborhood by taking advantage of the assumption of continuity of $\phi$.

Given the previously stated properties of the mappings, we can generally write, for any rigid displacement $\delta$, that:
\begin{align}
\begin{split}
\forall \epsilon, \epsilon' = \epsilon + \delta, \;\;\;\;
& \mathbf{s}_i = \mathbf{s}_{i'}
\\
\eqref{eq:phi} \Leftrightarrow
\; & \phi_{\epsilon}(\mathbf{m}_i) = \phi_{\epsilon'}(\mathbf{m}_{i'})
\\
\eqref{eq:phi} \Leftrightarrow 
\; & g\big(f(\mathbf{m}_i), \epsilon\big) = g\big(f(\mathbf{m}_{i'}), \epsilon'\big)
\\
\eqref{eq:f} \Leftrightarrow 
\; & g(\mathbf{p}_i, \epsilon) = g(\mathbf{p}_{i'}, \epsilon + \delta)
\\
\eqref{eq:delta} \Leftrightarrow 
\; & g(\mathbf{p}_i, \epsilon) = g(\mathbf{p}_{i'} - \delta, \epsilon)
\\
\eqref{eq:ge} \Leftrightarrow 
\; & g_\epsilon(\mathbf{p}_i) = g_\epsilon(\mathbf{p}_{i'} - \delta)
\\
\eqref{eq:bijection} \Leftrightarrow
\; & \mathbf{p}_{i} = \mathbf{p}_{i'} - \delta.
\end{split}
\label{eq:general_relation}
\end{align}
If both sensorimotor experiences $(\mathbf{m_i}, \mathbf{s_i})$ and $(\mathbf{m_{i'}}, \mathbf{s_{i'}})$ are collected in the same environmental state $\epsilon$, then $\delta = 0$ and we have:
\begin{equation}
\forall \epsilon,\:
\mathbf{s}_i = \mathbf{s}_{i'}
\Leftrightarrow
\phi_{\epsilon}(\mathbf{m}_i) = \phi_{\epsilon}(\mathbf{m}_{i'})
\Leftrightarrow
\mathbf{p}_{i} = \mathbf{p}_{i'}.
\label{eq:inv1}
\end{equation}
This relation is invariant to the environmental state $\epsilon$.
Through the sensory experiences collected when exploring environments, the agent can thus discover that different motor states are associated with the same external sensor position.
The necessary condition for such a discovery is that the agent can explore more than one sensorimotor pair $(\mathbf{m_i}, \mathbf{s_i})$ for any environmental state $\epsilon$.

If we now consider not two but four sensorimotor pairs such that two of them $\{(\mathbf{m_i}, \mathbf{s_i}), (\mathbf{m_j}, \mathbf{s_j})\}$ are collected in a first environmental state $\epsilon$, and the two others $\{(\mathbf{m_{i'}}, \mathbf{s_{i'}}), (\mathbf{m_{j'}}, \mathbf{s_{j'}})\}$ are collected in a second environmental state $\epsilon'$, we can write:
\begin{align}
\begin{split}
\forall \epsilon, \epsilon' = \epsilon + \delta,
& \begin{cases}
	\mathbf{s}_i = \mathbf{s}_{i'}\\
	\mathbf{s}_j = \mathbf{s}_{j'}
	\end{cases}
\\
\eqref{eq:phi}
\Leftrightarrow
& \begin{cases}
	\phi_{\epsilon}(\mathbf{m}_i) = \phi_{\epsilon'}(\mathbf{m}_{i'})\\
	\phi_{\epsilon}(\mathbf{m}_j) = \phi_{\epsilon'}(\mathbf{m}_{j'})
	\end{cases}
\\
\eqref{eq:general_relation}
\Leftrightarrow
& \begin{cases}
	\mathbf{p}_{i} = \mathbf{p}_{i'} - \delta\\
	\mathbf{p}_{j} = \mathbf{p}_{j'} - \delta
	\end{cases}
\\
\Leftrightarrow
& \;\;\; \mathbf{p}_{j} - \mathbf{p}_{i} = \mathbf{p}_{j'} - \mathbf{p}_{i'}.
\end{split}
\label{eq:inv2}
\end{align}
This relation is once again invariant to the environmental states $\epsilon$ and $\epsilon'$, as long as $\delta$ corresponds to a rigid displacement in space.
Through the sensory experience collected when exploring environments that can move, the agent can thus discover that different motor changes are associated with equivalent external displacements of the sensor.
To be precise,
%$\mathbf{m}_{i}\to\mathbf{m}_{i'}$ and $\mathbf{m}_{j}\to\mathbf{m}_{j'}$ are associated with the same external displacement $\overrightarrow{\mathbf{p}_{i}\mathbf{p}_{i'}} = \overrightarrow{\mathbf{p}_{j}\mathbf{p}_{j'}}$, and
$\mathbf{m}_{i}\to\mathbf{m}_{j}$ and $\mathbf{m}_{i'}\to\mathbf{m}_{j'}$ are associated with the same external displacement $\overrightarrow{\mathbf{p}_{i}\mathbf{p}_{j}} = \overrightarrow{\mathbf{p}_{i'}\mathbf{p}_{j'}}$.
The necessary condition for such a discovery is that the agent explores more than one sensorimotor pair $(\mathbf{m_i}, \mathbf{s_i})$ in more than one environmental state $\epsilon$.

\begin{figure}[t!]
\centering
\includegraphics[width=1\linewidth]{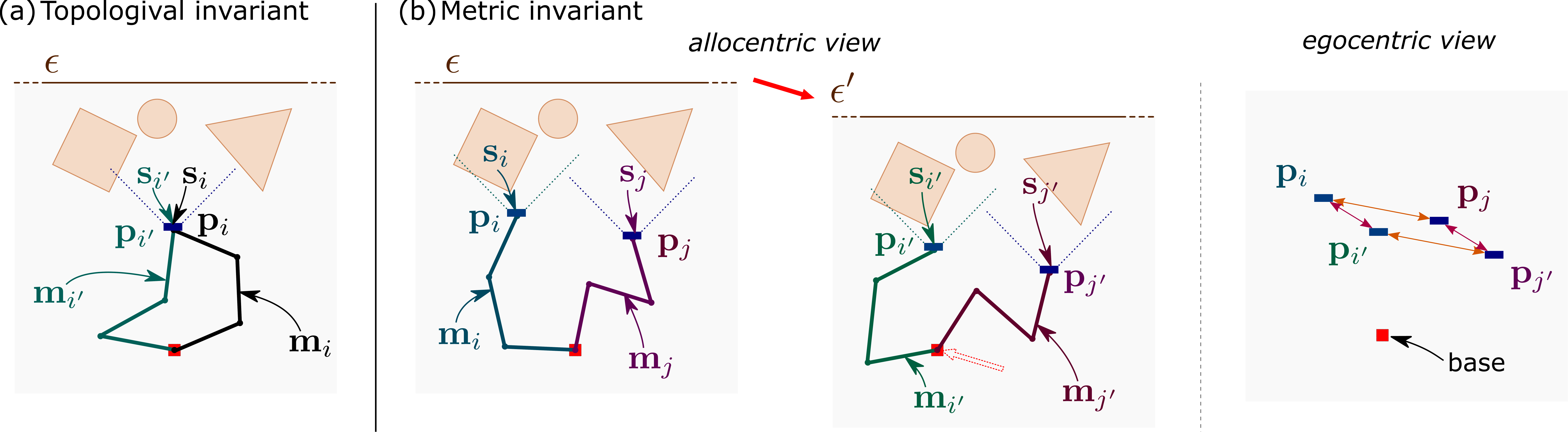}
\caption{Diagram illustrating topological and metric invariants with a three-segment arm agent. (a) Two motor configurations $\mathbf{m}_i$ and $\mathbf{m}_{i'}$ generating similar sensory states $\mathbf{s}_{i}$ and $\mathbf{s}_{i'}$ are associated with similar egocentric positions $\mathbf{p}_{i}$ and $\mathbf{p}_{i'}$ of the sensor.
(b) A displacement $\epsilon \rightarrow \epsilon'$ (red arrow) of the environment relative to the agent's base (red square) can be \textit{compensated} by the agent. As a result, two motor states $\mathbf{m}_{i}$ and $\mathbf{m}_{i'}$ are associated with similar sensory states $\mathbf{s}_{i}$ and $\mathbf{s}_{i'}$ before and after the displacement. The same logic applies to $\mathbf{m}_{j}$ and $\mathbf{m}_{j'}$, and their similar sensory states $\mathbf{s}_{j}$ and $\mathbf{s}_{j'}$. This means that the 4 corresponding egocentric sensor positions $\mathbf{p}_{i}$, $\mathbf{p}_{i'}$, $\mathbf{p}_{j'}$, and $\mathbf{p}_{j}$ form a parallelogram in space. It is represented on the right, relatively to the agent's base, with colored double-headed arrows indicating metric equivalences of displacements between its vertices.
Note that a displacement of the environment is equivalent to a displacement of the base of the agent (red dotted arrow).
(Best seen in color)}
\label{fig:diagram_topology_metric}
\end{figure}

Finally, let's re-express equations \eqref{eq:inv1} and \eqref{eq:inv2} as follows:
\begin{equation}
\forall \epsilon,\:
|\phi_{\epsilon}(\mathbf{m}_i) - \phi_{\epsilon}(\mathbf{m}_{i'})| = 0
\Leftrightarrow
|\mathbf{p}_{i} - \mathbf{p}_{i'}| = 0,
\label{eq:inv1_}
\end{equation}
\begin{align}
\begin{split}
\forall \epsilon, \epsilon' = \epsilon + \delta,
& \begin{cases}
	|\phi_{\epsilon}(\mathbf{m}_i) - \phi_{\epsilon'}(\mathbf{m}_{i'})| = 0\\
	|\phi_{\epsilon}(\mathbf{m}_j) - \phi_{\epsilon'}(\mathbf{m}_{j'})| = 0
	\end{cases}
\\
\Leftrightarrow
& \;\;\; | (\mathbf{p}_{j} - \mathbf{p}_{i}) - (\mathbf{p}_{j'} - \mathbf{p}_{i'}) | = 0,
\end{split}
\label{eq:inv2_}
\end{align}
where $|.|$ denotes a norm, and let's take advantage of the continuity of the mappings $f$, $g$, and $\phi$, to generalize them to local neighborhoods:
\begin{equation}
\forall \epsilon,\:
|\phi_{\epsilon}(\mathbf{m}_i) - \phi_{\epsilon}(\mathbf{m}_{i'})| \ll \mu
\Leftrightarrow
|\mathbf{p}_{i} - \mathbf{p}_{i'}| \ll \mu,
\label{eq:inv1_final}
\end{equation}
\begin{align}
\begin{split}
\forall \epsilon, \epsilon' = \epsilon + \delta,
& \begin{cases}
	|\phi_{\epsilon}(\mathbf{m}_i) - \phi_{\epsilon'}(\mathbf{m}_{i'})| \ll \mu\\
	|\phi_{\epsilon}(\mathbf{m}_j) - \phi_{\epsilon'}(\mathbf{m}_{j'})| \ll \mu
	\end{cases}
\\
\Leftrightarrow
& \;\;\; | (\mathbf{p}_{j} - \mathbf{p}_{i}) - (\mathbf{p}_{j'} - \mathbf{p}_{i'}) | \ll \mu,
\end{split}
\label{eq:inv2_final}
\end{align}
where $\mu$ is a small value.
\\
Due to their nature, we refer to the invariants of Equation~\eqref{eq:inv1_final} as \emph{topological invariants}, and the ones of Equation~\eqref{eq:inv2_final} as \emph{metric invariants}.
A simple illustration of both types of invariants is proposed in Fig.~\ref{fig:diagram_topology_metric}.

\section{Neural network architecture and training}
\label{sec:Neural network and training}

\subsection{Neural network}

We propose a simple neural network architecture to perform sensorimotor prediction.
It is composed of two types of module: $\text{Net}_\text{enc}$ and $\text{Net}_\text{pres}$.
\\
The $\text{Net}_\text{enc}$ module projects a motor state $\mathbf{m}_t$ onto a representation $\mathbf{h}_t$ of dimension $N_h$.
It consists of a fully connected Multi-Layer Perceptron (MLP) with three hidden layers of respective sizes $(150,100,50)$ with SeLu activation functions \cite{klambauer2017self}, and a final output layer of size $N_h$ with linear activation functions.
This last layer corresponds to the space $\mathbb{R}^{N_h}$ in which the motor representation are analyzed in this work.
\\
The $\text{Net}_\text{pred}$ module takes as input the concatenation $(\mathbf{h}_{t}, \mathbf{h}_{t+1}, \mathbf{s}_{t})$ of a current motor representation, a future motor representation, and a current sensory state, and outputs a prediction $\mathbf{\tilde{s}}_{t+1}$ of the future sensory state $\mathbf{s}_{t+1}$.
It consists of a fully connected MLP with three hidden layers of respective sizes $(200,150,100)$ with SeLu activation functions, and a final output layer of size $N_s$ with linear activation functions.
\\
The overall network architecture connects the predictive module $\text{Net}_\text{pred}$ to two siamese copies of the $\text{Net}_\text{enc}$ module, ensuring that both motor states $\mathbf{m}_t$ and $\mathbf{m}_{t+1}$ are consistently encoded using the same mapping.
\\
Note that the simulations have also be run with ReLu units \cite{nair2010rectified} in place of the SeLu units and produced qualitatively identical results.

\begin{figure}[t!]
\centering
\includegraphics[width=0.75\linewidth]{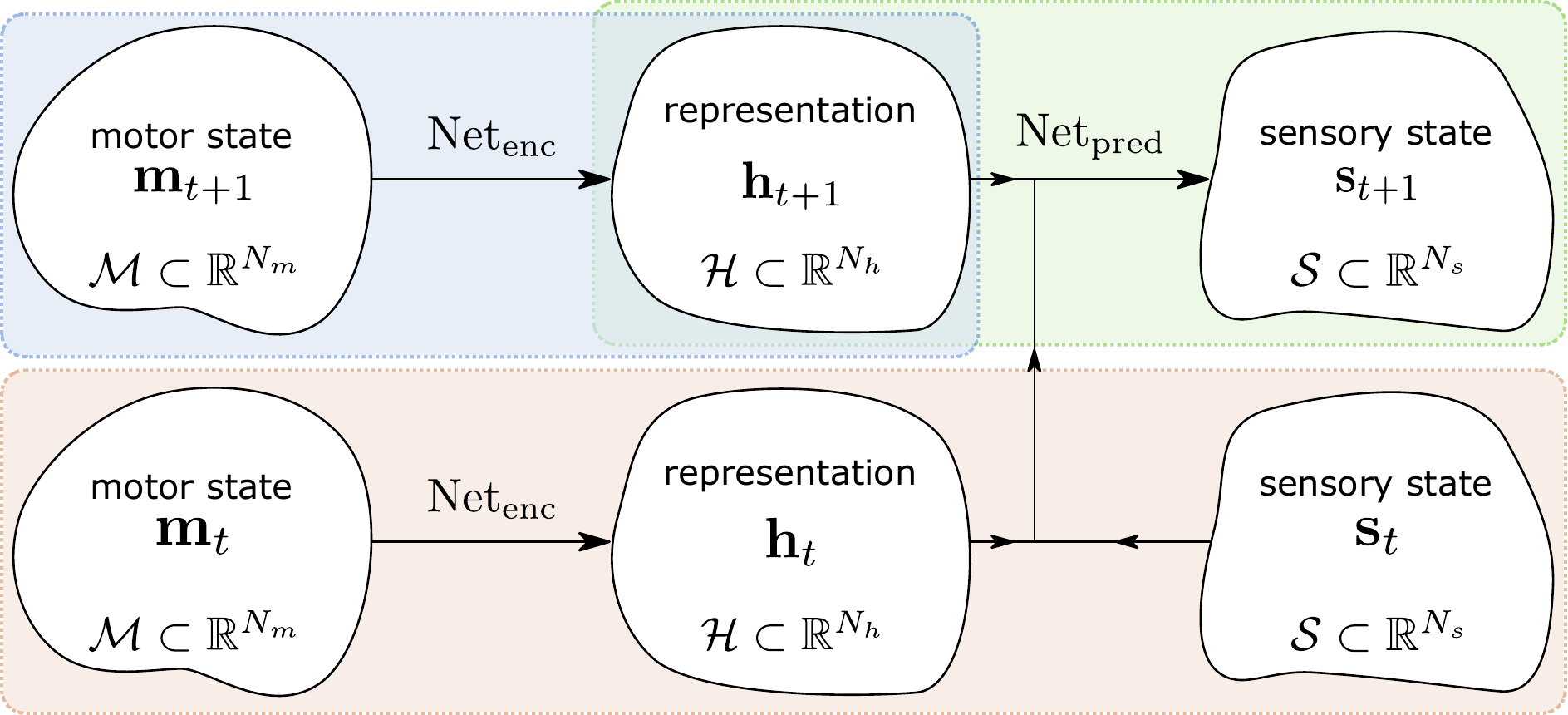}
\caption{Illustration of the variables processed by the neural architecture. (Best seen in color)}
\label{fig:mapping_network}
\end{figure}

As illustrated in Fig.~\ref{fig:mapping_network}, a parallel can be drawn between the network illustrated in Fig.\ref{fig:setups} and the sensorimotor mapping illustrated in Fig.~\ref{fig:mapping_functions}.
Indeed, one can relate $\text{Net}_\text{enc}$ to the forward mapping $f$, and $\text{Net}_\text{pred}$ to the sensory mapping $g$.
However, as $\epsilon$ is not directly accessible to the agent, its counterpart in the network corresponds to the current sensorimotor pair, where $\mathbf{m}_{t}$ is encoded as $\mathbf{h}_t$ (red frame in Fig.~\ref{fig:mapping_network}). As we assume the sensorimotor experience to be unambiguous, the sensorimotor experience $(\mathbf{m}_{t}, \mathbf{s}_{t})$ indeed carries information about the state of the environment $\epsilon$.
\\
Note that in agreement with the hypothesis of Sec.~\ref{sec:Mathematical formalism} and the fact that $\text{Net}_\text{enc}$ instantiates a continuous mapping, the set $\mathcal{H} \subset \mathbb{R}^{N_h}$ of all representations $\mathbf{h}$ is also assumed to be a manifold.

\subsection{Loss minimization}

The unsupervised (or self-supervised) objective of the network is to minimize the MSE loss:
$$
\text{Loss} = \frac{1}{K} \sum_{k=1}^K |\mathbf{\tilde{s}}_{t+1}^{(k)} - \mathbf{s}_{t+1}^{(k)}|^2,
$$
where $k$ denotes a sample index, $K$ is the number of samples in the training dataset, and $|.|$ denotes the Euclidean norm.
Importantly, no particular component is added to the loss function regarding the structure of the representation $\mathbf{h}$ built by the network.
The loss is minimized using the ADAM optimizer \cite{kingma2014adam}, with a learning rate linearly decreasing from $10^{-3}$ to $10^{-5}$ in $8\times10^{4}$ epochs, and a mini-batch size of $100$ sensorimotor transitions.
The optimization is stopped after $10^{5}$ epochs (a single mini-batch is fed to the network at each epoch).

\subsection{Training data}

The training data are generated by having the simulated agent explore its environment and collect sensorimotor transitions $(\mathbf{m}_t, \mathbf{s}_t) \rightarrow (\mathbf{m}_{t+1}, \mathbf{s}_{t+1})$.
A total of $150000$ transitions are collected for each simulation by randomly sampling pairs $(\mathbf{m}_t, \mathbf{m}_{t+1})$ in the motor space (uniform distribution) and collecting the corresponding sensory inputs.
Depending on the type of exploration, the environment also translates during the data collection, instantaneously  changing its position in the horizontal plane relative to the agent's base:

\textit{MEM case:} for each transition, the environment translates between the collection of $(\mathbf{m}_t, \mathbf{s}_t)$ and $(\mathbf{m}_{t+1}, \mathbf{s}_{t+1})$.
This ensures that the sensorimotor experiences collected by the agent do not fulfill condition I, and a fortiori condition II.

\textit{MM case:} the environment never translates and keeps its initial position during the collection of all sensorimotor transitions.
This ensures that the sensorimotor experiences collected by the agent fulfill condition I, but not condition II.

\textit{MME case:} for each transition, the environment translates after the collection of both $(\mathbf{m}_t, \mathbf{s}_t)$ and $(\mathbf{m}_{t+1}, \mathbf{s}_{t+1})$.
This ensures that the sensorimotor experiences collected by the agent fulfill conditions I and II.

%Finally, an additional transformation has been implemented in the MM case to mitigate the additional constraints it induces on the sensory experience. Indeed, as the environment never moves and the agent's working space is smaller than the entire room, the sensory distribution experienced by the agent is overall different (less variability) in the MM case than in the two other types of exploration. Moreover, it is possible for multiple obstacles to be located close to the center of the room, which greatly limits the number of viable sensor positions. As a consequence, the motor representation built by the network can be greatly skewed in the vanilla M case.
%In order to mitigate these effects, the length of the arm segments are doubled in the MM case in order for the sensor to cover a larger part of the environment during exploration, and to reduce the effect of objects blocking a large part of the working space.
%This transformation is not necessary, but leads to a fairer comparison between the different types of exploration.
%Note however that the problem of the obstacle blocking some parts of the working space is reduced but still remains. It is thus impossible for the agent to get some sensory information for some parts of its motor space. This is not a problem in the MEM and MME case, as obstacles move relatively to the agent during exploration.

Before being fed to the network, the whole collected dataset is normalized such that each motor and sensory component spans $[-1, 1]$ over the whole dataset (performed independently for each simulation).
For each training epoch, $100$ quadruplets $(\mathbf{m}_t, \mathbf{s}_{t}, \mathbf{m}_{t+1}, \mathbf{s}_{t+1})$ are randomly drawn in the dataset to form a mini-batch.

\subsection{Remarks}

The overall neural network architecture and training procedure have been kept simple. No particular heuristics have been added to improve convergence, generalization, or any other property of the network such as sparsity. Similarly, the architecture's meta-parameters have not been optimized beyond simply checking that the network was expressive enough to approximate the expected mappings. The same size of $\text{Net}_\text{enc}$ and $\text{Net}_\text{pred}$ have for instance been used in both simulations, even if the sensorimotor mapping is significantly more complex in the Arm in a room simulation than in the Discrete world simulation.
Moreover, the network's generalization capacity has not been evaluated.
\\
This is because the prime goal of this work is not to optimize a neural network to efficiently solve a task, but rather to study if spatial invariants are captured as a byproduct of sensorimotor prediction, without the need for additional priors.

The code to generate the sensorimotor data, train the neural network, and analyze the motor representation is available at \url{https://github.com/alaflaquiere/learn-spatial-structure}. % XXXX  URL TO ADD

\section{Simulations}
\label{sec:Simulations}

\begin{figure*}[t!]
\centering
\includegraphics[width=0.90\linewidth]{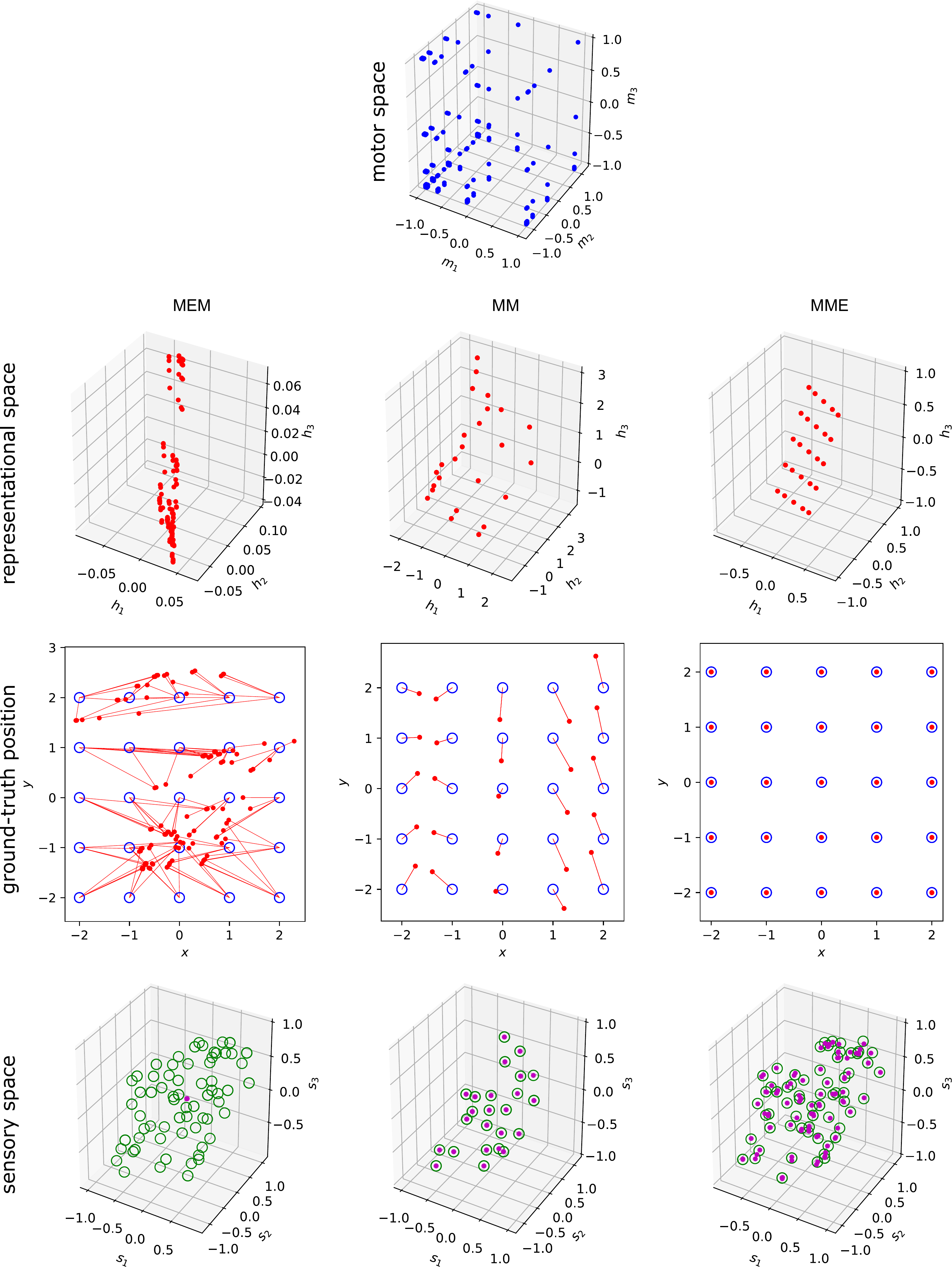}
\caption{Visualization of the normalized regular motor sampling $\mathbf{m}$ (blue dots), its representation $\mathbf{h}$ in the representational space (red dots), and the corresponding ground truth position $\mathbf{p}$ (blue circles) for the three types of exploration in the Discrete world simulation.
The affine projection $\mathbf{h}^{(p)}$ of the representations are also displayed in the space of positions. Lines have been added to visualize the distances between each $\mathbf{h}^{(p)}$ and its ground truth counterpart $\mathbf{p}$.
Finally, the predicted sensory states $\tilde{\mathbf{s}}_{t+1}$ (magenta dots) outputted by the network are displayed in the 3 first dimensions of the sensory space, alongside the ground-truth sensory states $\mathbf{s}_{t+1}$ (green circles).
(Best seen in color)}
\label{fig:full_projection_discreteworld}
\end{figure*}

Two different agent-environment systems are simulated to generate sensorimotor experiences:

\textit{Discrete world}:
This corresponds to an artificial setup designed to optimally evaluate the impact of the experience of sensorimotor invariants on the motor representation built by the network (no sensory ambiguity, continuous sensorimotor mapping, no border effect).
\\
The environment consists in a grid world of size $10 \times 10$.
Each square of the grid is associated with a sensory state $\mathbf{s}$ of dimension $N_s = 4$ that a sensor can capture.
This sensory state is set to vary smoothly with the position $(r, c)$ of the square in the grid.
To ensure such a smoothness, each component $s_i$ of the sensory state $\mathbf{s} = [s_1, s_2, s_3, s_4]$ is defined as a sum of random periodic functions varying with respect to $r$ or $c$:
\begin{equation}
s_i =
\sum_k^3 \frac{1}{\lambda_{1,k}^i} \cos \bigg( 2 \pi \Big(\lfloor \lambda_{1,k}^i \rceil \frac{r}{10} + \lambda_{2,k}^i \Big) \bigg)
+ \frac{1}{\lambda_{3,k}^i} \cos \bigg( 2 \pi \Big(\lfloor \lambda_{1,k}^i \rceil \frac{c}{10} + \lambda_{4,k}^i \Big) \bigg),
\end{equation}
where all $\lambda$ parameters are randomly drawn in $[-2, 2]$, and $\lfloor . \rceil$ denotes the rounding operation necessary to ensure that the frequency of the function is a multiple of the size of the grid.
\\
The agent's base (invisible for the sensor) can be placed in any square of the grid.
The agent has a sensor that it can move in a working space of size $5 \times 5$ centered on its base.
In each square, the sensor receives the corresponding sensory state $\mathbf{s}$.
To change its sensor position, the agent generates motor states $\mathbf{m} = [m_1, m_2, m_3]$ of dimension $N_m = 3$.
Each motor state $\mathbf{m}$ is associated with an egocentric position $\mathbf{p}$ of the sensor in the working space in the following way:
\begin{equation}
\mathbf{p} = 4 \times [\sqrt[3]{m_1}, \sqrt[3]{m_2}] - 2,
\end{equation}
where each $m_i$ lives in $[0, 1]$.
This arbitrary forward mapping is purposefully made non-linear and redundant, as $m_3$ does not affect the sensor position.
Because of this non-linearity and the discrete nature of the grid world, the agent can only sample its motor space in a non-linear fashion (see Fig.~\ref{fig:full_projection_discreteworld}).
Note that this forward mapping is artificial, and we did not define any actual physical body to instantiate it. 
\\
Finally, the environment can translate rigidly with respect to the agent's base, effectively moving the working space in the whole grid.
The amplitude of this translation is drawn uniformly in $[-10, 10]$ for both its horizontal and vertical components.
The whole grid world is set to act as a torus, which means that the sensor appears on the other side of the grid displayed in Fig.~\ref{fig:setups} when the working space extends beyond its limits.

\begin{figure*}[t!]
\centering
\includegraphics[width=0.96\linewidth]{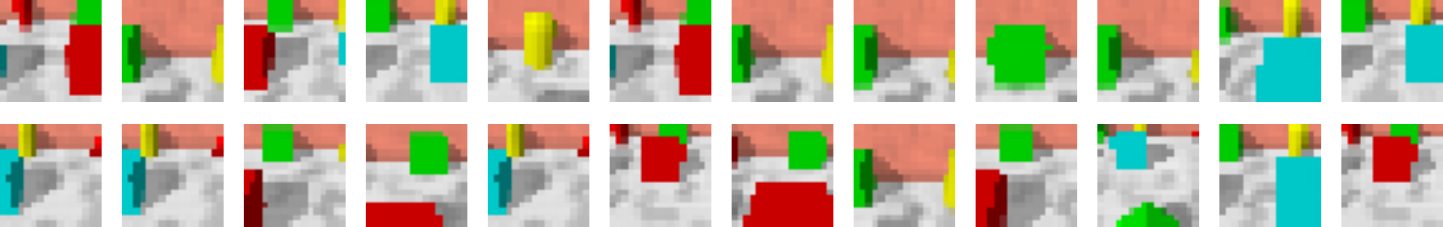}
\caption{Examples of sensory states received by the agent in the Arm in a room simulation. The spatial and RGB structure of each image is discarded before being fed to the network by flattening its $16 \times 16 \times 3$ values into a simple vector of length $768$. (Best seen in color)}
\label{fig:room_samples}
\end{figure*}

\textit{Arm in a room}:
This corresponds to a more complex and realistic setup in which an arm explores a 3D room.
The environment is similar to the one proposed in \cite{eslami2018neural}.
The room is of size $7 \times 7$ units, has walls, and is filled with $16$ random simple geometric objects.
The textures and colors of the walls/floor and objects are picked randomly at the beginning of the simulation.
The objects are distributed along a regular grid, but disturbed with an additional small displacement drawn in $U(-0.3, 0.3)^2$ in order to add some randomness.
\\
The agent is a three-segment arm moving in the horizontal plane at a height of 1.6 units.
It is equipped with a RGB camera of resolution $16\times16$ at its end, orientated with a downward tilt of $0.62$ rad, and generating a sensory state $\mathbf{s}$ of dimension $N_s=16\times16\times3=768$. Note that due to its orientation, the sensor can see the objects in the room but not the arm segments (see Fig.~\ref{fig:room_samples}).
The three arm segments are each of length 0.5 unit.
Their respective relative orientation are controlled in $[-\pi,\pi]$ radians by three independent components of the motor state $\mathbf{m}$ of dimension $N_m=3$.
The effective working space of the agent is thus an horizontal disk of diameter $3$ units.
During the arm movements, the orientation of the sensor is kept fixed.
\\
Finally, the environment can translate rigidly with respect to the arm's base with a maximal horizontal and vertical range of $[-1.75, 1.75]$, where a translation of $[0, 0]$ corresponds to the room being centered on the agent's base. The cumulative effect of the agent's movements and the environment's movements is such that the sensor never moves outside the walls of the room.

\section{Detailed results analysis}
\label{sec:Detailed results analysis}

The results of Fig.~\ref{fig:results} and Fig.~\ref{fig:projection} are here analyzed in more details.

\subsection{Discrete world}

\paragraph{MEM exploration:}
One can see in Fig.~\ref{fig:results} that the loss stays relatively high during the whole training. This is due to the fact that it is impossible to accurately predict the future sensory input $\mathbf{s}_{t+1}$ as the environment is always moving between $t$ and $t+1$.
As a consequence, the network learns to output the average sensory state which minimizes the MSE (see the sensory space in Fig.~\ref{fig:full_projection_discreteworld}).

The topological dissimilarity $D_{topo}$ also stays at a relatively high value during training, and displays an important standard deviation.
For each run, $D_{topo}$ varies greatly during the whole training, although it tends to stabilize after $8 \times 10^4$ epochs, when the learning rate reaches its minimum value.
This behavior seems to indicate that the topology of $\mathbf{h}$ differs from the one of $\mathbf{p}$, and that the network tends to build an arbitrary motor representation.
This is confirmed in Fig.~\ref{fig:full_projection_discreteworld}, where both $\mathbf{h}$ and its affine projection $\mathbf{h}^{(p)}$ in the space of positions display an arbitrary topology compared to the one of the ground-truth $\mathbf{p}$.

The metric dissimilarity $D_{metric}$ also stays at a relatively high value during training, and displays an important standard deviation.
Once again, $D_{metric}$ varies greatly for each run, but stabilizes a bit after $8 \times 10^4$ epochs.
Just like for the topology, this seems to indicate that the metric of $\mathbf{h}$ differs from the one of $\mathbf{p}$, which is confirmed in Fig.~\ref{fig:full_projection_discreteworld}.

\paragraph{MM exploration:}
One can see in Fig.~\ref{fig:results} that the loss quickly converges to very small values. It is expected, as a static environment ensures that the network can easily learn to map the motor states to their corresponding sensory states.
This is confirmed in Fig.~\ref{fig:full_projection_discreteworld} where the future sensory states appear to be accurately predicted.

The topological dissimilarity $D_{topo}$ also quickly converges to very small values.
This seems to indicate that the topology of $\mathbf{h}$ is similar to the one of $\mathbf{p}$.
This is confirmed in Fig.~\ref{fig:full_projection_discreteworld}, where one can see that all redundant motor states associated with the same sensor position are encoded with the same representation, and that the global topology of the manifold of representations is equivalent to the one of the ground-truth position. The manifold of motor encoding is thus practically of dimension $2$, when the motor space is actually of dimension $3$.

The metric dissimilarity $D_{metric}$ converges to an average value, between $0$ and its value in the MEM case. Its standard deviation is also significant.
This seems to indicate that the metric of $\mathbf{h}$ differs from the one of $\mathbf{p}$, which is confirmed in Fig.~\ref{fig:full_projection_discreteworld}.
This lower value of $D_{metric}$ compared to the MEM case is however due to the fact that capturing the topology of $\mathbf{p}$ in $\mathbf{h}$ necessarily entails that the metric difference between the two is lower than compared to a random projection.

\paragraph{MME exploration:}
Just like in the MM case, the loss quickly converges to very small values as the consistency of the sensorimotor transitions ensures that the network can predict the future sensory state based on the current sensorimotor pair.
This is once again confirmed in Fig.~\ref{fig:full_projection_discreteworld} where the future sensory states are accurately predicted by the network.

The topological dissimilarity $D_{topo}$ also quickly converges to very small values.
This seems to indicate that the topology of $\mathbf{h}$ is similar to the one of $\mathbf{p}$.
This is confirmed in Fig.~\ref{fig:full_projection_discreteworld}, where one can see that all redundant motor states associated with the same sensor position are encoded with the same representation, and that the global topology of the manifold of representations is equivalent to the one of the ground-truth position.

Finally, the metric dissimilarity $D_{metric}$ converges to a very small value, with a very small standard deviation.
This seems to indicate that the metric of $\mathbf{h}$ is similar to the one of $\mathbf{p}$, which is confirmed in Fig.~\ref{fig:full_projection_discreteworld}. In particular, we can see that $\mathbf{h}^{(p)}$ perfectly aligns with the grid of ground-truth positions. Thus there exists a simple affine transformation between the motor representation built by the network and the external position of the sensor.

\subsection{Arm in a room}
\label{sec:differences}

\begin{figure*}[t!]
\centering
\includegraphics[width=0.9\linewidth]{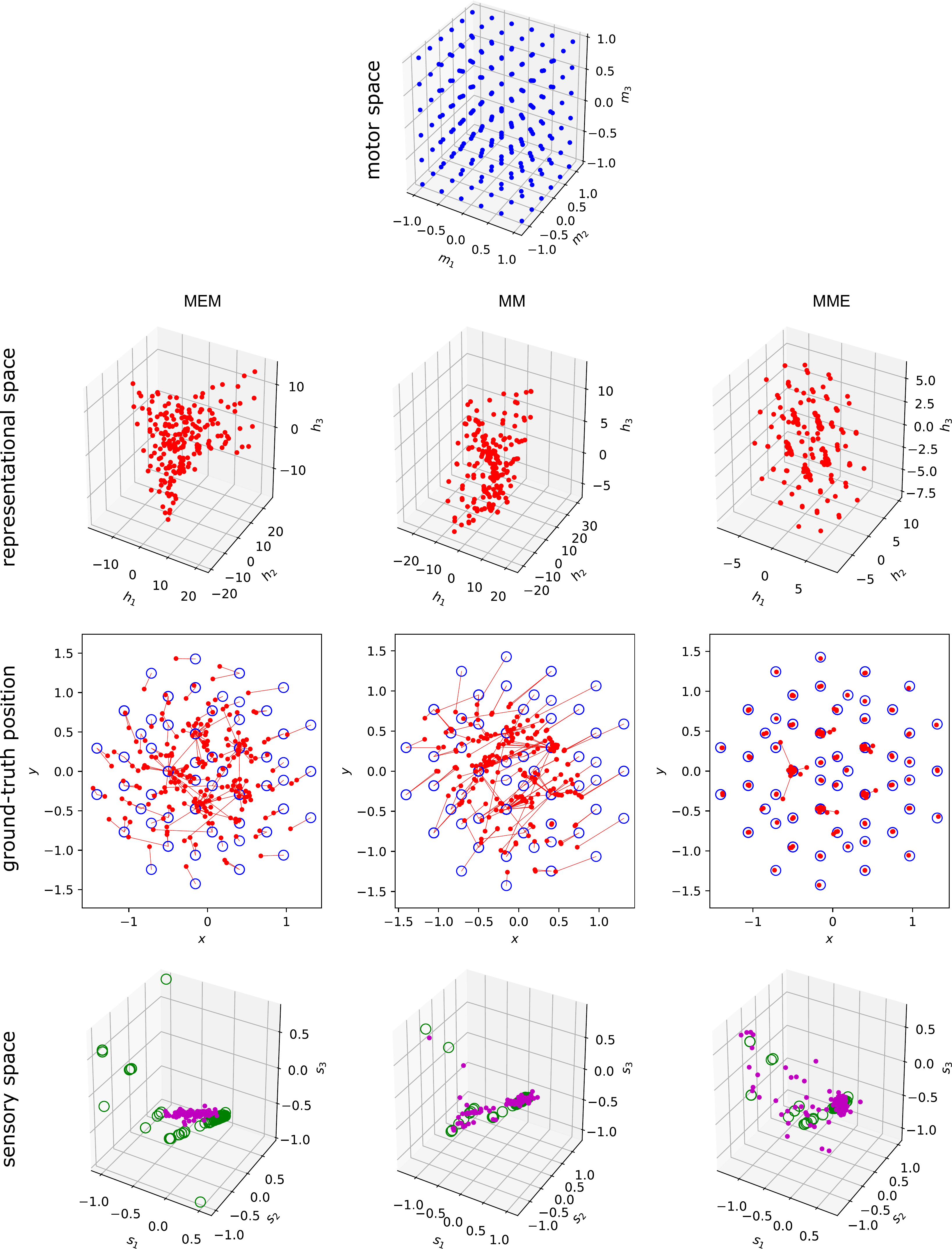}
\caption{Visualization of the normalized regular motor sampling $\mathbf{m}$ (blue dots), its representation $\mathbf{h}$ in the representational space (red dots), and the corresponding ground truth position $\mathbf{p}$ (blue circles) for the three types of exploration in the Arm in a room simulation.
The affine projection $\mathbf{h}^{(p)}$ of the representations are also displayed in the space of positions. Lines have been added to visualize the distances between each $\mathbf{h}^{(p)}$ and its ground truth counterpart $\mathbf{p}$.
Finally, the predicted sensory states $\tilde{\mathbf{s}}_{t+1}$ (magenta dots) outputted by the network are displayed in the 3 first dimensions of the sensory space, alongside the ground-truth sensory states (green circles).
(Best seen in color)}
\label{fig:full_projection_armroom}
\end{figure*}

The analysis done for the Discrete world also globally applies to the Arm in a room simulation. We thus only focus on the differences below.

\paragraph{Expressivity of $\text{Net}_{\text{pred}}$:}
The sensorimotor mapping to learn is significantly more complex in the Arm in a room simulation, and the sensory space is of significantly higher dimension.
As a consequence the neural network, and in particular the $\text{Net}_{\text{pred}}$ module, is not expressive enough to perfectly predict future sensory states, even in cases when it should be theoretically possible (MM and MME cases).
This results in the loss and its standard deviation being of greater amplitude than in the Discrete world simulation.
This phenomenon is also illustrated in Fig.~\ref{fig:full_projection_armroom}, as one can see in the sensory space that the predictions outputted by the network do not perfectly match the ground-truth in the MM and MME cases.
\\
Note that this lack of expressivity could be seen as a shortcoming of our evaluation. But on the contrary we consider it as a highlight, as it shows that the motor representation tends to capture topological and metric invariants even when the overall performance of the sensorimotor prediction is limited.

\paragraph{Limited exploration:}
An important difference between the Discrete world and Arm in a room simulations is that the exploration is limited in the latter.
Indeed, in the former, the amplitude of environmental translations, coupled with the fact that the grid world acts as a torus, ensures that any motor state can be associated with the sensor being in any square of the grid.
As a consequence, all motor states are statistically associated with the same distribution of sensory states over the whole environment.
On the contrary, in the Arm in a room simulation, the environment is limited by walls.
As a consequence, each motor state only covers a sub-part of the environment when the latter moves. For instance, a motor state corresponding to the arm being extended to the left means that the sensor will never experience the sensory states on the far right of the room. Thus each motor state is associated with a slightly different sensory distribution when the environment moves.
It is thus possible for the network to infer an approximation of the topology of $\mathbf{p}$ which helps to reduce the MSE, even in the MEM case.
This can be seen in Fig.~\ref{fig:full_projection_armroom} as the motor representation tends to capture the topology of $\mathbf{p}$ in the MEM case. This effect is also visible in the sensory space as one can see that, in this simulation, the network does not simply output the same average sensory state for all motor states. Instead, each motor state is associated with the average sensory state over the slightly different sensory distribution is its associated with.
\\
The limited exploration thus impacts the measures $D_{topo}$ and $D_{metric}$ in the MEM case. They are lower than what could be expected if we simply extrapolated from the results of the previous simulation.

\paragraph{Ambiguous environments:}
The third difference is that the 3D room environments can present some sensory ambiguities: very different sensor positions can be associated with very similar sensory states.
As a consequence, the sensory manifold associated with the manifold of positions can be twisted in a non-trivial way in the sensory space. In case of perfect ambiguity between two (or more) positions, the topology of the manifold even changes locally.
This perturbs the learning of the motor representation when the environment is static (MM case), and leads to $D_{topo}$ measures greater than what could be expected if we simply extrapolated from the results of the previous simulation.
Note however that this sensory ambiguity is not a problem in the MEM case, as the movements of the environment ensure that the different sensor positions which are ambiguous for a given environment position are associated with different sensory distributions over the whole exploration.

\begin{figure}[t!]
\centering
\includegraphics[width=1\linewidth]{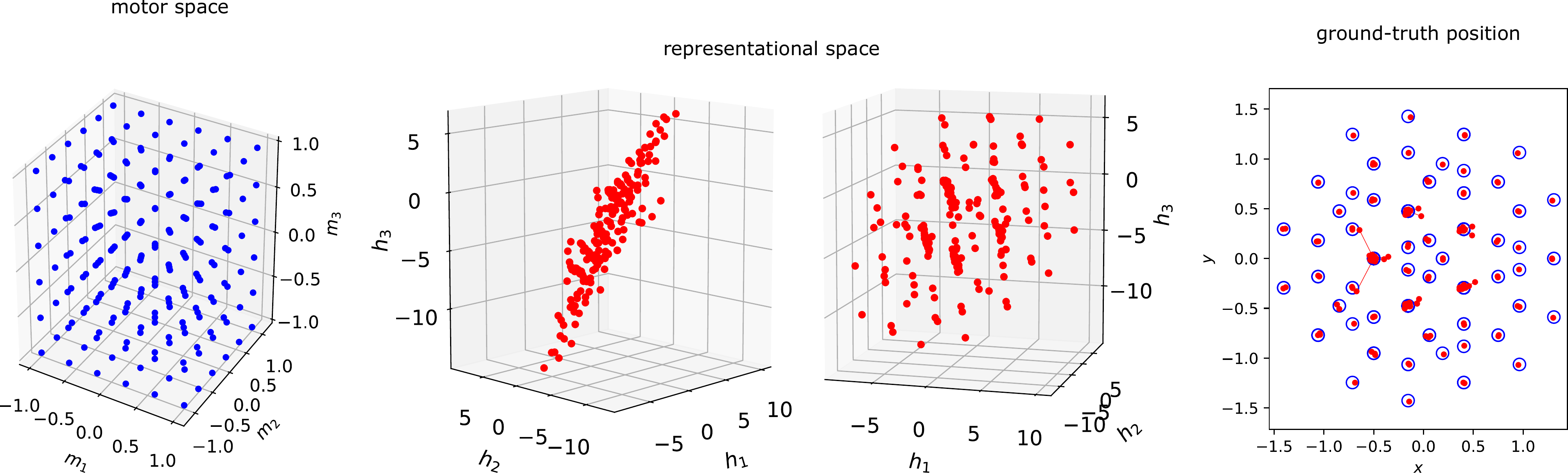}
\caption{Visualization of the normalized regular motor sampling $\mathbf{m}$ (blue dots), its representation $\mathbf{h}$ in the representational space (red dots), and the corresponding ground truth position $\mathbf{p}$ (blue circles) for an Arm in a room simulation and an MME exploration. The affine projections $\mathbf{h^{(p)}}$ are also displayed in the space of positions. In this trial, the manifold of $\mathbf{h}$ slightly spread along a third dimension in the representational space. (Best seen in color)}
\label{fig:non-flat}
\end{figure}

\paragraph{Non-flat representations:} 
The final difference is the one that leads to the standard deviation of $D_{topo}$ being more important than expected in the MME case.
As can be observed in Fig~\ref{fig:non-flat}, it sometimes happens in the MME case that the manifold of $\mathbf{h}$ appears slightly spread in the 3D representational space instead of approximating a 2D flat manifold.
Yet, the affine projection of $\mathbf{h}$ in the space of $\mathbf{p}$ properly aligns with the ground-truth positions.
This phenomenon is due to the way a neural network processes data.
Indeed the representation $\mathbf{h}$ is fed to a fully connected layer where each neuron performs a linear projection of $\mathbf{h}$ before passing it through its activation function.
As a consequence, the network has two equivalent options to respect the sensory invariants induced by the MME exploration:
i) flatten (to 2D) the manifold in the 3D representational space, or
ii) tune the weights to the next fully connected layer such that all neurons perform projections which take into account only two dimensions in the representational space.
In both cases, the input received by the predictive module $\text{Net}_\text{enc}$ is equivalent.
This also explains why even a non-flat manifold of $\mathbf{h}$ still matches the ground-truth position, as the linear regression correspond to a projection of the same nature as the one implemented by the connections to the next layer.
The effect of such a phenomenon can also be seen in Fig.~\ref{fig:results}, as it explains why $D_{topo}$ shows a significant standard deviation in the MME case.

\section{Additional experiments}
\label{sec:Additional experiments}

\subsection{Representational space of higher dimension}
\label{sec:Representational space of higher dimension}

\begin{figure}[t!]
\centering
\includegraphics[width=1\linewidth]{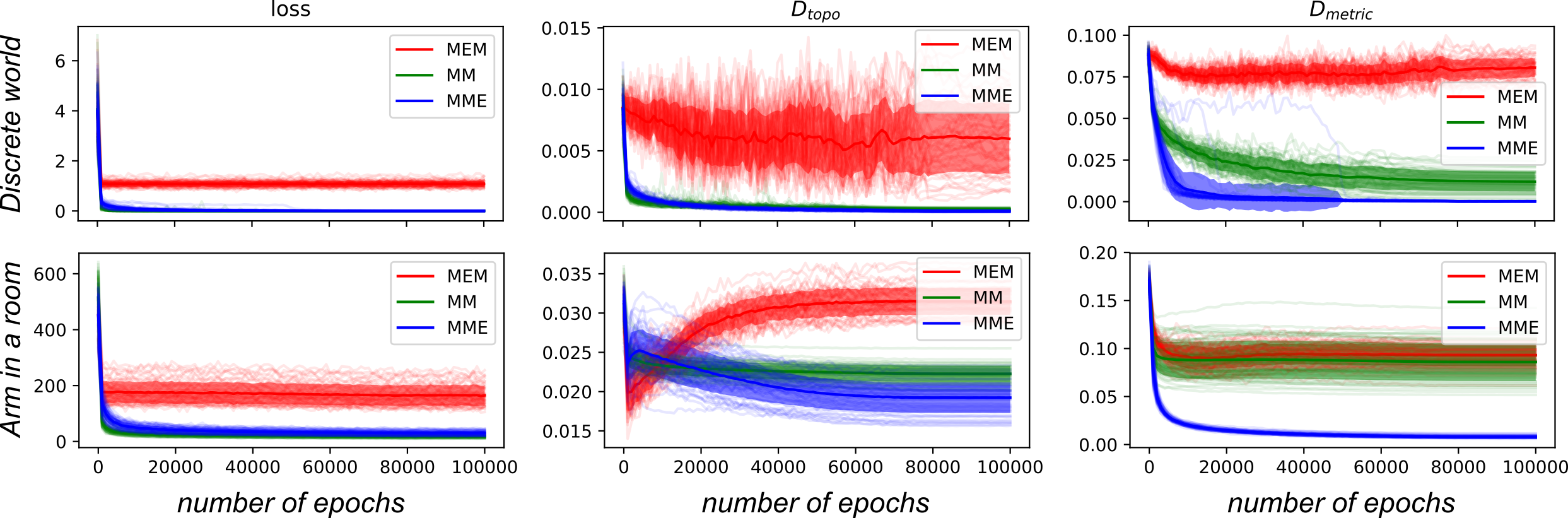}
\caption{Evolution of the loss and the dissimilarity measures $D_{topo}$ and $D_{metric}$ during training for both setups, for the three types of exploration, and with $N_h=25$ instead of $3$. The displayed means and standard deviations are computed over 50 independent runs. (Best seen in color)}
\label{fig:bigger_h}
\end{figure}

The main results were obtained with a representational space $\mathbb{R}^{N_h}$ of dimension $N_h = 3$ to facilitate visualization.
The same exact experiments were also run with a representational space of dimension $N_h = 25$. The results, presented in Fig.~\ref{fig:bigger_h}, are qualitatively equivalent to the ones described previously\footnote{The higher dimension also implies an even bigger effect of the potential twist of the sensory manifold and of the spread of the representation in more than 2 dimensions (see Sec~\ref{sec:differences}).}. This seems to indicate that the dimension of the representational space has no influence on the way space-induced invariants are captured in $\mathbf{h}$.

\subsection{Forward mappings of higher complexity}
\label{sec:Forward mappings of higher complexity}

The robustness of the results with respect to the complexity of the agent's body has been evaluated by designing more complex forward mappings.
In the Discrete world setup, we designed a new agent with $N_m=6$ motors and the following forward mapping:
\begin{equation}
\mathbf{p} = \begin{bmatrix} x \\ y \end{bmatrix} = 4 \times (P \cdot A \cdot f(\mathbf{m})) - 2
\text{,\hspace{0.1cm} with: }
f\left(
\begin{bmatrix}
m_1 \\ m_2 \\ m_3 \\ m_4 \\ m_5 \\ m_6
\end{bmatrix}
\right)
=
\begin{bmatrix}
m_1^2 \\
\sqrt{m_2} \\
\sqrt[3]{m_3} \\
0.1 \times \left(\frac{1.1}{0.1}\right)^{m_4} - 0.1 \\
\log(m_5 \times (e^1 - 1) + 1 ) \\
m_6
\end{bmatrix},
\end{equation}
where $A$ is a $6\times6$ mixing matrix with random elements uniformly drawn in $[-2,2]$, and $P$ is a diagonal projection matrix whose first two elements are equal to $1$ and the others to $0$.
The elements of $\mathbf{m}$ are sampled from $[0,1]$ and mapped to $[0,1]$ by the non-linear transformation $f$. The random matrix $A$ is also carefully designed such that the mixed components of $A \cdot f(\mathbf{m})$ still belong to $[0,1]$. Finally, a simple linear transformation is applied after the projection $P$ such that the sensor coordinates $x$ and $y$ lie in $[-2, 2]$.
Intuitively, the overall transformation consists in passing the 6 motor components through some non-linearities, mixing them via a random matrix $A$, and finally projecting the result in 2D via the matrix $P$.
\\
In the Arm in a room setup, we designed a four-segment arm agent with $N_m=6$ motor:, four hinge joints, and two translational joints on the central two segments.
The arm still moves in the horizontal plane, but the (maximum) length of its segments has been reduced to 0.375 so that the working space's radius is unchanged. The hinge joints are identical to the previous agent, and translational joints are controlled in [-0.375, 0.375].
\\
Due to the increased complexity of the forward mapping to estimate, we increased the size of the $\text{Net}_{\text{enc}}$ module to $(500, 400, 300, 200)$ (in both setups), and increased the number of collected exploratory transitions to $300000$ instead of $150000$ (in the Arm in a room setup).

\begin{figure}[h!]
\centering
\includegraphics[width=1\linewidth]{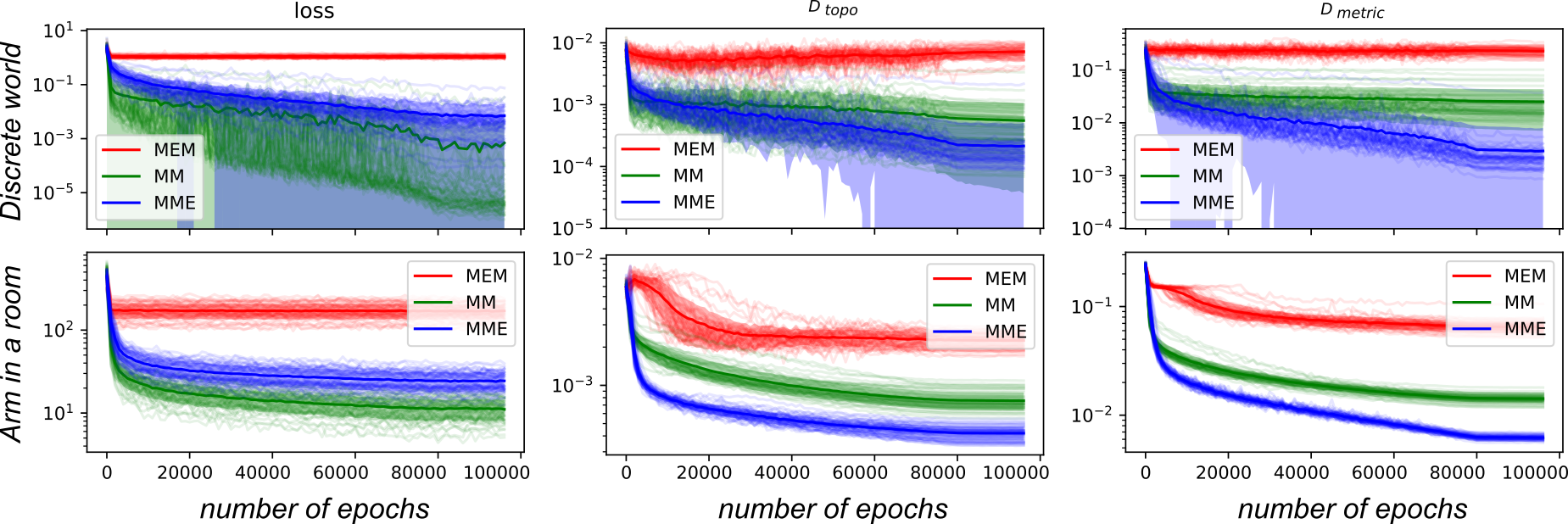}
\caption{Evolution of the loss and the dissimilarity measures $D_{topo}$ and $D_{metric}$ during training for both setups, for the three types of exploration, and with the more complex agents (6 degrees of freedom). A log-scale is used on the y-axis due to the large values induced by the MEM exploration. The displayed means and standard deviations are computed over 50 independent runs. (Best seen in color)}
\label{fig:manydof}
\end{figure}

\begin{figure*}[t!]
\centering
\includegraphics[width=0.9\linewidth]{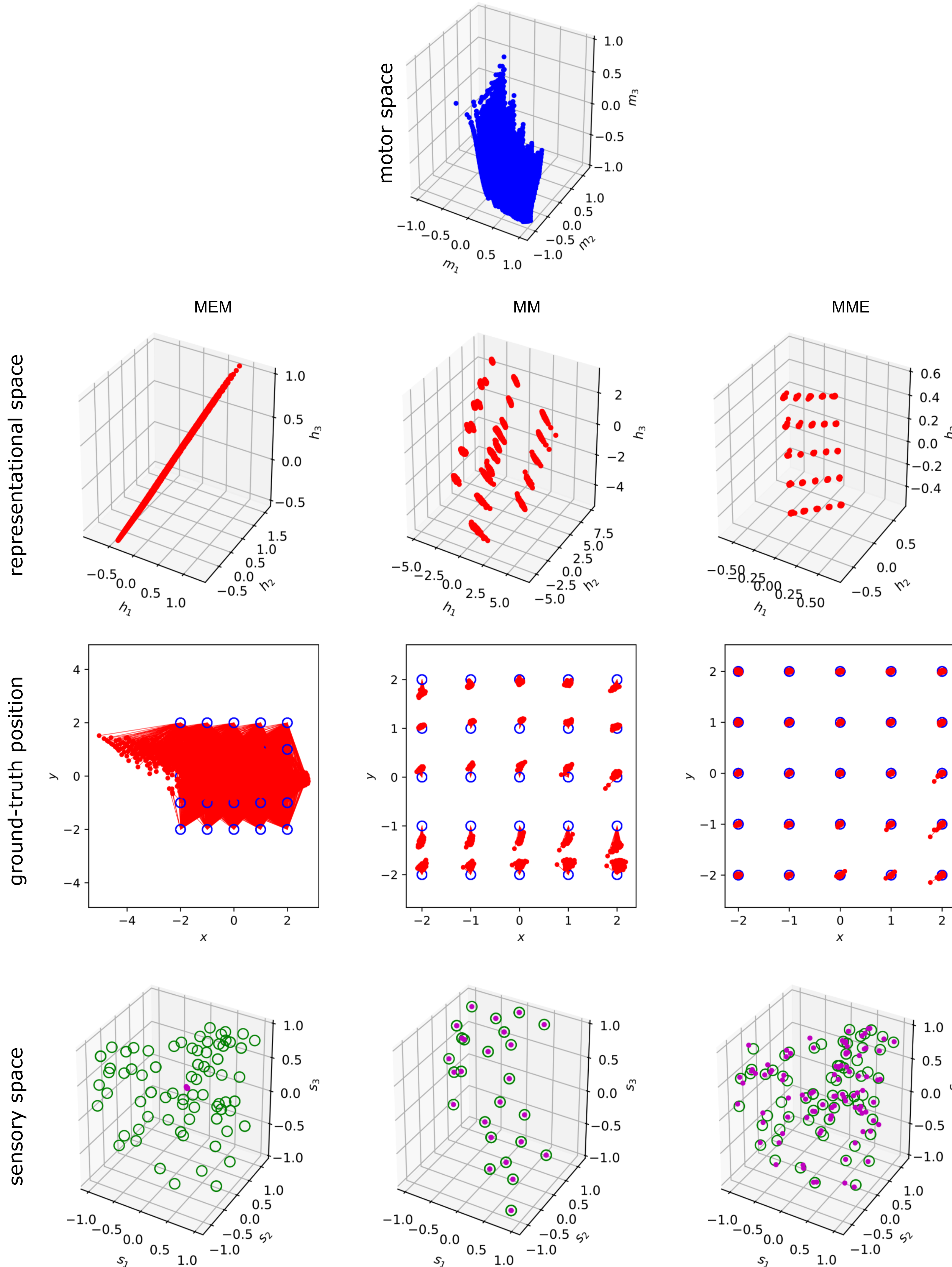}
\caption{Visualization of the normalized regular motor sampling $\mathbf{m}$ (blue dots) in the 3 first dimensions of the 6-D motor space, its representation $\mathbf{h}$ in the representational space (red dots), and the corresponding ground truth position $\mathbf{p}$ (blue circles) for the three types of exploration in the Discrete world simulation with the more complex agent.
The affine projection $\mathbf{h}^{(p)}$ of the representations are also displayed in the space of positions. Lines have been added to visualize the distances between each $\mathbf{h}^{(p)}$ and its ground truth counterpart $\mathbf{p}$.
Finally, the predicted sensory states $\tilde{\mathbf{s}}_{t+1}$ (magenta dots) outputted by the network are displayed in the 3 first dimensions of the sensory space, alongside the ground-truth sensory states (green circles).
(Best seen in color)}
\label{fig:full_projection_manydofgridexplorer}
\end{figure*}

\begin{figure*}[t!]
\centering
\includegraphics[width=0.9\linewidth]{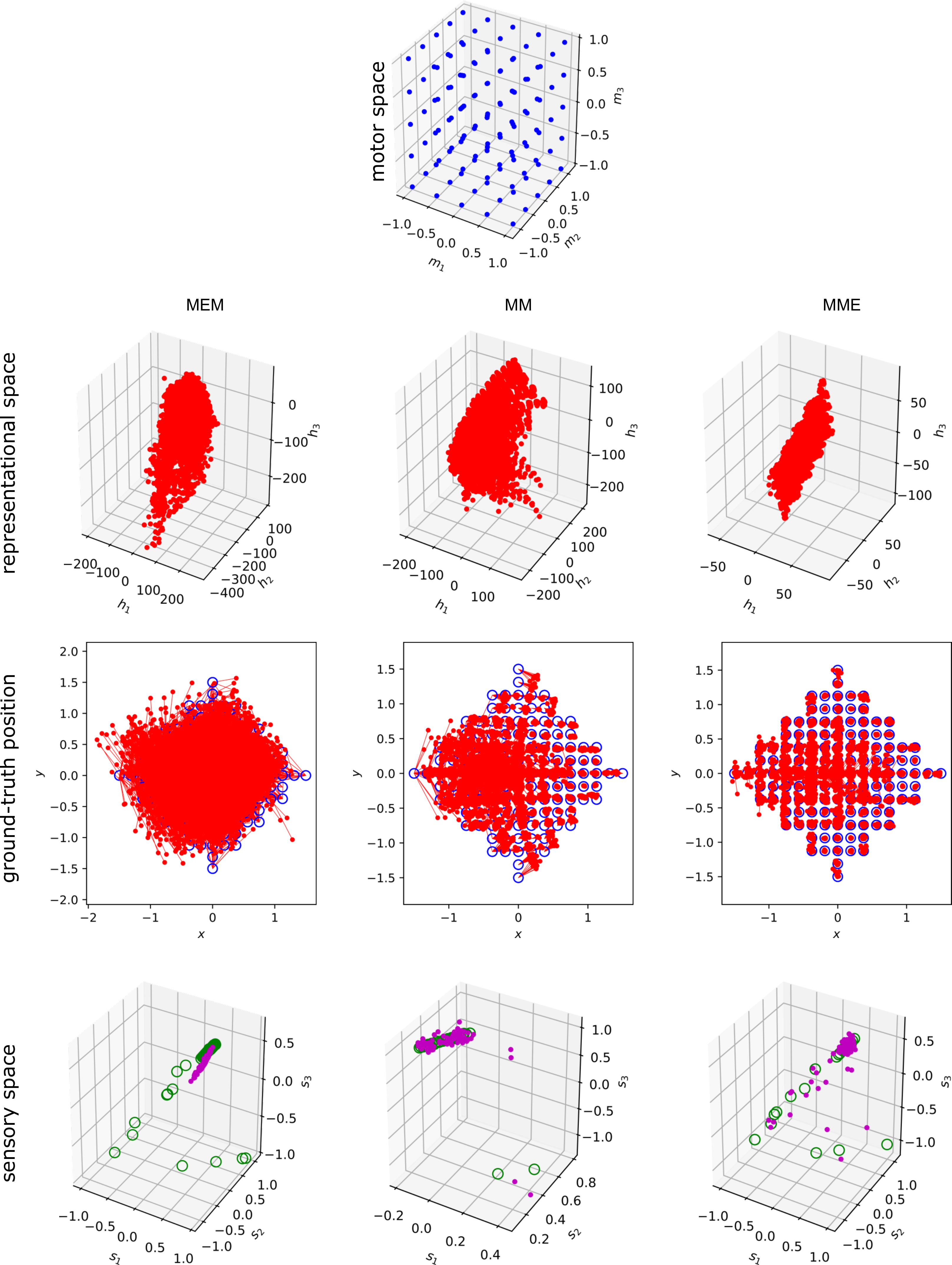}
\caption{Visualization of the normalized regular motor sampling $\mathbf{m}$ (blue dots) in the 3 first dimensions of the 6-D motor space, its representation $\mathbf{h}$ in the representational space (red dots), and the corresponding ground truth position $\mathbf{p}$ (blue circles) for the three types of exploration in the Arm in a room simulation with the more complex agent.
The affine projection $\mathbf{h}^{(p)}$ of the representations are also displayed in the space of positions. Lines have been added to visualize the distances between each $\mathbf{h}^{(p)}$ and its ground truth counterpart $\mathbf{p}$.
Finally, the predicted sensory states $\tilde{\mathbf{s}}_{t+1}$ (magenta dots) outputted by the network are displayed in the 3 first dimensions of the sensory space, alongside the ground-truth sensory states (green circles).
(Best seen in color)}
\label{fig:full_projection_manydofarmroom}
\end{figure*}

The results of the experiments with these more complex forward mappings are presented in Figs.~\ref{fig:manydof}, \ref{fig:full_projection_manydofgridexplorer}, and \ref{fig:full_projection_manydofarmroom}.
They are qualitatively equivalent to the ones observed with the original forward models. This seems to indicate that the complexity of the forward model (body of the agent) has no influence on the way space-induced invariants are captured in $\mathbf{h}$.
\\
We can however note some quantitative difference in Fig.~\ref{fig:manydof}. Indeed, due to the increased dimension of the motor space, the regular sampling performed during the network evaluation requires significantly more motor samples ($15625$ instead of $125$ and $216$ in the original Discrete world and Arm in a room setups respectively). The loss is thus computed over more samples.
The higher number of degrees of freedom also means a higher degree of redundancy. As a result, more motor states are associated with the same egocentric position of the sensor. This has an indirect effect on the values observed for $D_{topo}$ and $D_{metric}$, as the arbitrary encoding induced by the MEM exploration results in significantly higher dissimilarity measures.
A log scale has then been used in Fig.~\ref{fig:manydof} in order to better visualize the results.

\subsection{Intrinsic stochasticity of the training}
\label{sec:Intrinsic stochasticity of the training}

\begin{figure}[h!]
\centering
\includegraphics[width=1\linewidth]{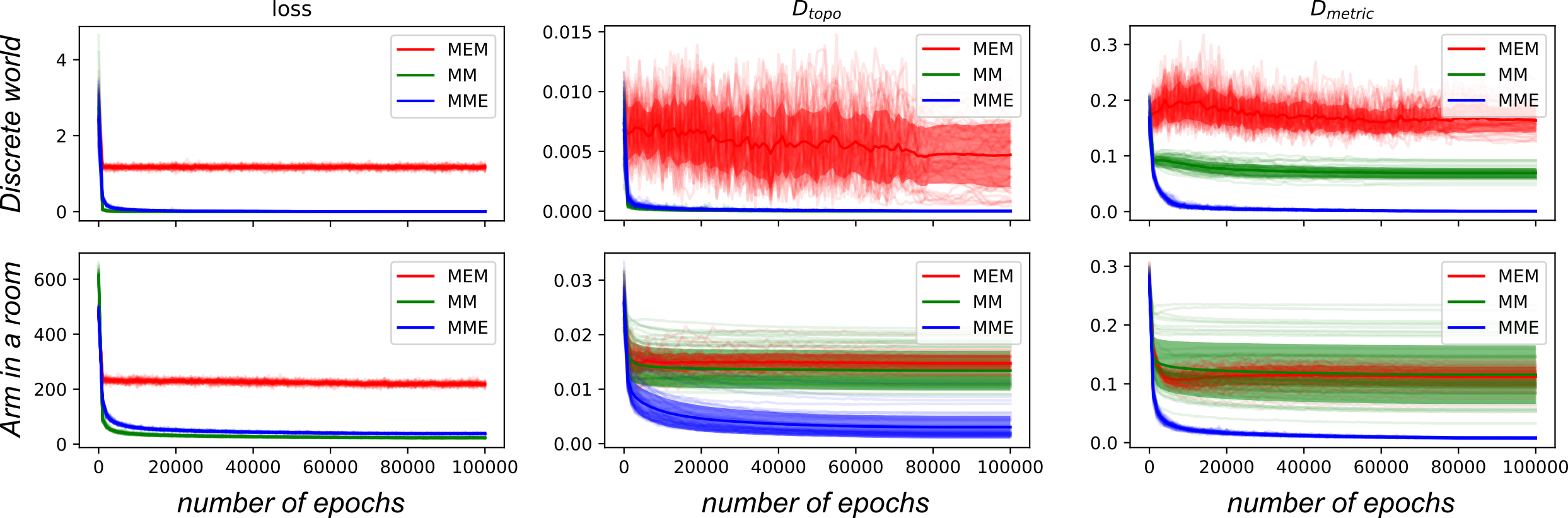}
\caption{Evolution of the loss and the dissimilarity measures $D_{topo}$ and $D_{metric}$ during training for both setups, for the three types of exploration, and with $N_h=3$. The displayed means and standard deviations are computed over 50 independent runs trained on a single dataset for each simulation. (Best seen in color)}
\label{fig:intrinsic_variability}
\end{figure}

Due to the stochasticity of the training procedure (network initialization, mini-batches selection), the learning curves display some intrinsic variability, even when trained on a fixed dataset.
We estimated this variability and display in Fig.~\ref{fig:intrinsic_variability} the average and standard deviation associated with $50$ independent runs trained on the same dataset, for each simulation.
\\
Compared to Fig.~\ref{fig:results}, one can see that the resulting standard deviations are very similar. This seems to indicate that most of the variability observed in the results is due to the intrinsic variability of the training procedure, rather than to the datasets on which the networks are trained.

\end{document}